%% file: template.tex
\documentclass{article}
\usepackage{threeparttable}   
\usepackage{diagbox}          
\usepackage{booktabs}

\usepackage{arxiv}
\usepackage{algorithm}        
\usepackage{algpseudocode}    
\usepackage[utf8]{inputenc} 
\usepackage[T1]{fontenc}    
\usepackage{hyperref}       
\usepackage{url}            
\usepackage{booktabs}       
\usepackage{amsfonts}       
\usepackage{nicefrac}       
\usepackage{microtype}      
\usepackage{lipsum}         
\usepackage{graphicx}
\usepackage{amsmath}
\usepackage{cleveref}       
\usepackage{multirow}
\usepackage[numbers]{natbib}
\usepackage{graphicx}
\usepackage{doi}

\title{QoSGMAA: A Robust Multi-Order Graph Attention and Adversarial Framework for Sparse QoS Prediction}

\newif\ifuniqueAffiliation
\uniqueAffiliationtrue  

\ifuniqueAffiliation
\author{
    \textbf{Guanchen Du} \\
    College of Mathematics and Computer Science\\
    Shantou University, Shantou, Guangdong, China \\
    \texttt{22gcdu@stu.edu.cn} \\
    \And
    \textbf{Jianlong Xu} \thanks{Corresponding author.} \\
    College of Mathematics and Computer Science\\
    Shantou University, Shantou, Guangdong, China \\
    \texttt{xujianlong@stu.edu.cn} \\
    \And
    \textbf{Mingtong Li} \\
    College of Mathematics and Computer Science\\
    Shantou University, Shantou, Guangdong, China \\
    \texttt{22mtli@stu.edu.cn} \\
    \And
    \textbf{Ruiqi Wang} \\
    College of Mathematics and Computer Science\\
    Shantou University, Shantou, Guangdong, China \\
    \texttt{22rqwang@stu.edu.cn} \\
    \And
    \textbf{Quanqing Guo} \\
    College of Mathematics and Computer Science\\
    Shantou University, Shantou, Guangdong, China \\
    \texttt{20qqguo@stu.edu.cn} \\
    \And
    \textbf{Caiyi Chen} \\
    College of Mathematics and Computer Science\\
    Shantou University, Shantou, Guangdong, China \\
    \texttt{21cychen@stu.edu.cn} \\
    \And
    \textbf{Qingcao Dai} \\
    College of Mathematics and Computer Science\\
    Shantou University, Shantou, Guangdong, China \\
    \texttt{21qcdai@stu.edu.cn} \\
    \And
    \textbf{Yuxiang Zeng} \\
    College of Computer Science and Electronic Engineering\\
    Hunan University, Changsha, Hunan, China \\
    \texttt{zengyuxiang@hnu.edu.cn} \\
}
\fi

\date{\textbf{Under review in \textit{ACM Transactions on Internet Technology (TOIT)}}}

\renewcommand{\shorttitle}{}

\hypersetup{
pdftitle={QoSGMAA: A Robust Multi-Order Graph Attention and Adversarial Framework for Sparse QoS Prediction},
pdfsubject={cs.ML},
pdfauthor={Guanchen Du et al.},
pdfkeywords={Quality of Service prediction, Web services, Graph neural networks, Multi-order attention mechanism, Adversarial learning, Gumbel-Softmax sampling, Sparse matrix modeling, User-service interaction, Deep learning for service computing},
}

\begin{document}
\maketitle

\begin{abstract}
With the rapid advancement of internet technologies, network services have become critical for delivering diverse and reliable applications to users. However, the exponential growth in the number of available services has resulted in many similar offerings, posing significant challenges in selecting optimal services. Predicting Quality of Service (QoS) accurately thus becomes a fundamental prerequisite for ensuring reliability and user satisfaction.
However, existing QoS prediction methods often fail to capture rich contextual information and exhibit poor performance under extreme data sparsity and structural noise. To bridge this gap, we propose a novel architecture, QoSMGAA, specifically designed to enhance prediction accuracy in complex and noisy network service environments. QoSMGAA integrates a multi-order attention mechanism to aggregate extensive contextual data and predict missing QoS values effectively. Additionally, our method incorporates adversarial neural networks to perform autoregressive supervised learning based on transformed interaction matrices. To capture complex, higher-order interactions among users and services, we employ a discrete sampling technique leveraging the Gumbel-Softmax method to generate informative negative samples. Comprehensive experimental validation conducted on large-scale real-world datasets demonstrates that our proposed model significantly outperforms existing baseline methods, highlighting its strong potential for practical deployment in service selection and recommendation scenarios.
\end{abstract}

\keywords{Quality of Service prediction, Web services, Graph neural networks, Multi-order attention mechanism, Adversarial learning, Gumbel-Softmax sampling, Sparse matrix modeling, User-service interaction, Deep learning for service computing}

\section{Introduction}
\input{text/1Introduction}

\section{Related Work}

\input{text/2RelatedWork}
\section{Question Definition and Framework Overview}
\input{text/3FQDQ}

\section{Methodology}
\input{text/4Solution}

\section{Experinment}
\input{text/5Exp}

\section{Conclusion and Future Work}
\input{text/6Con}

\section*{Acknowledgments}
This research was financially supported by Guangdong Basic and Applied Basic Research Foundation (No.2023A1515010707, 2024A1515012468), Central Guiding Local Science and Technology Development Special Fund Project (No. STKJ2024083), Special Projects in Key Fields of Guangdong Universities (No. 2022ZDZX1008), Guangdong Province Special Fund for Science and Technology ("major special projects + task list") Project (No. STKJ202209017), and Guangdong Science and Technology Plan (No.STKJ2023012).

\bibliographystyle{ieeetr}
\bibliography{references} 







\end{document}

%% file: text/1Introduction.tex
Quality of Service (QoS) is a critical non-functional attribute in the Web service domain. Its primary performance metrics include response time, throughput, and similar indicators \cite{zeng2003quality}.
 QoS prediction, as an essential research task in the field of service computing, aims to model the interaction between the user and the service to achieve accurate prediction of the QoS performance of target services. With the rapid development of cloud computing and Internet of Things (IoT) technologies, Web services with similar functionalities have experienced exponential growth. The absence of effective QoS evaluation mechanisms often results in suboptimal service selection, thereby significantly degrading user experience \cite{liang2020secure}. It is worth noting that high-precision QoS prediction models play an important supporting role for the downstream service ecosystem, and their prediction results can effectively serve key application scenarios such as service selection\cite{peng2025energy}, service composition\cite{wu2024constraint}, and service recommendation\cite{cao2024prkg}. Therefore, how to construct an efficient and reliable QoS prediction mechanism has become a fundamental scientific problem to be solved in the field of service computing \cite{liu2019context}.

 A foundational and early recognized method for QoS prediction is Collaborative Filtering (CF). This approach primarily utilizes historical QoS data from users, leveraging similarities among users to predict unknown values \cite{zheng2020web}. While CF-based methods effectively capture latent patterns in user-service interactions, they suffer from limitations such as cold-start problems and high computational complexity \cite{zeng2023gatcf}. These constraints often impede their applicability in dynamic and sparse data environments typical of many real-world scenarios.

In recent years, deep learning technologies have demonstrated remarkable success in domains such as computer vision (CV) \cite{chen2024enhancing}, natural language processing (NLP) \cite{lin2023evolutionary}, and time-series prediction \cite{du7time}. The current research has also evolved to include graph neural networks (GNNs) for QoS prediction \cite{li2020graphmf}. GNNs significantly enhance the effectiveness of embedding matrices by capturing the high-order relationships between users and services. Although GNNs have been widely adopted to model high-order dependencies \cite{jin2022graph}, existing approaches often struggle to effectively capture such relationships, particularly under severe matrix sparsity. Moreover, current methods\cite{liu2023qosgnn} demonstrate limited capability in handling interactions between learned user graphs and service graphs, limiting their practical utility in real-world applications.

 According to the previous research, we find that the current QoS prediction field still faces the following main issues:

\begin{enumerate}
    \item \textbf{Sparse Matrix}: In QoS prediction, the interactions between users and services often exhibit data sparsity, which significantly hampers the predictive model's ability to capture the potential relationships between users and services. Existing methods, such as Graph Attention Networks (GATs), do not effectively leverage node connections, particularly with multi-level neighbors. This limitation leads to difficulty accurately predicting unknown service quality values, especially under sparse conditions.
    \item \textbf{Ineffective interaction}: Current QoS prediction models fall short in integrating features between users and services, especially in their capability to handle noisy data. These models often rely on static feature representation methods, such as matrix factorization(MF), overlooking the dynamic changes in user-service relationships. Moreover, existing technologies usually struggle to distinguish useful information from noise effectively, affecting the model's generalization ability and prediction accuracy. 
\end{enumerate}

Capturing potential relationships between users and services has always been a significant challenge in QoS Prediction. To address this issue, we introduce a multi-order graph attention mechanism. Compared to traditional graph convolutional networks, this mechanism assigns different weights to neighbor features for each user (or service), enhancing the flexibility and accuracy of GNNs in handling node relationships. However, conventional graph attention mechanisms typically only perform weighted aggregation for first-order neighbors. In real applications, a node is often influenced by its neighbors' neighbors, i.e., more distant nodes. Therefore, we employ a multi-order graph attention mechanism, which allows interaction and integration of features across different layers and captures information from more distant nodes, providing richer context. Especially in extremely sparse graph data, multi-order graph attention effectively mitigates the difficulty of information capture by integrating multi-hop neighbor information.

Existing methods exhibit apparent shortcomings in modeling the interactions between users and services, particularly in handling noisy data in QoS prediction. Traditional methods mainly use matrix factorization techniques for feature integration, which attempt to learn embeddings but often struggle to effectively handle complex interactions between features. To address this, we adopt a fully connected neural network model with learnable weight parameters to facilitate these interactions, aiming to capture higher-order interaction relationships. Considering the potential presence of noise in the data, we introduce adversarial neural networks at the interaction stage to enhance the model's robustness and maintain performance stability when facing imperfect data. Unlike conventional continuous negative sampling(based on constant samples), we leverage a discrete Gumbel-Softmax sampling technique to efficiently generate negative samples, enabling gradient-based optimization while simulating realistic anomalies and providing more precise QoS predictions in complex data environments.

Specifically, our core contributions are as follows:

\begin{itemize}
    \item We design a multi-order graph attention neural network to capture potential relationships between users and services, enabling the model to learn and predict from a broader context.
    \item We introduce adversarial neural networks during the information interaction phase to enhance the model's adaptability and robustness in complex data environments.
    \item We implement a discrete sampling method based on Gumbel-Softmax in adversarial neural networks to generate negative samples for training, improving the model's generalization ability.
    \item Extensive experiments on large-scale real-world datasets demonstrate our framework's superior performance and robustness compared to state-of-the-art methods.
\end{itemize}

The remainder of the paper is organized as follows: Section 2 introduces related work. Section 3 provides an overview of the problem formulation and system model design. Section 4 details our design and complete solution. Section 5 evaluates the performance of our model under different settings. Section 6 concludes the paper and discusses future work.

%% file: text/2RelatedWork.tex
Accurate QoS prediction is essential for enabling various high-quality service provisioning strategies. Existing approaches predominantly rely on collaborative filtering, deep learning, and graph neural networks.

Early research on QoS prediction predominantly adopted Collaborative Filtering (CF) methods, including memory-based, model-based, and hybrid approaches. Memory-based methods estimate QoS values by calculating the similarity between users or services, leveraging data from the top-$k$ most similar neighbors, such as UPCC \cite{UPCC}, IPCC \cite{IPCC}, and UIPCC \cite{UIPCC}. However, such methods often suffer from data sparsity and face challenges in incorporating auxiliary contextual information. Model-based approaches, such as LNLFM \cite{yu2014personalized}, CloudPred \cite{zhang2011exploring}, AMF \cite{zhu2017online}, and CSMF \cite{wu2018collaborative}, predict unknown QoS values by learning latent factors of users and services. However, these methods still encounter convergence issues when applied to sparse data. For instance, Zou et al. \cite{zou2020ndmf} proposed the NDMF model, which enhances matrix factorization by integrating neighborhood information collaboratively selected by users and utilizing deep neural networks. This model incorporates both location context and historical call records to improve QoS prediction accuracy. However, the model exhibits poor performance in cold-start scenarios.

Hybrid methods combine the advantages of memory-based and model-based approaches to improve prediction accuracy. For instance, Huang et al. \cite{huang2021qos} proposed a deep learning-based model that integrates memory-based and model-based CF strategies to enhance QoS prediction in cloud services. While these combined methods typically yield superior results, they also tend to adopt the inherent drawbacks associated with the increasing complexity of the model and inherit the limitations of both memory-based and model-based techniques, typically relying on external information that may not be readily accessible.

With the rapid advancement of deep learning, these techniques have been widely adopted in QoS prediction. For example, Hussain et al. \cite{hussain2022assessing} introduced an IOWA (Induced Ordered Weighted Averaging) operation layer in the prediction layer to optimize the management and processing of large-scale QoS data. Yin et al. \cite{yin2020qos} proposed a novel QoS prediction method combining denoising autoencoders and fuzzy clustering techniques (DAFC). However, Xiang et al. \cite{zeng2023gatcf} noted that these methods remain ineffective under sparse data conditions, primarily due to their limited exploitation of graph structural information.

Graph Neural Networks (GNNs) are a powerful class of neural architectures with greater expressive capability than conventional neural networks \cite{jin2022graph}. GNNs operate via message passing, allowing each node to aggregate features from its neighbors. This mechanism naturally models interactions between users and services, making GNNs particularly effective for capturing complex relational patterns. For example, Long et al. \cite{li2020graphmf} combined GNNs with collaborative filtering to predict QoS in blockchain services using graph structure information. Liu et al. \cite{liu2023qosgnn} incorporated the Transformer architecture \cite{vaswani2017attention} into GNNs to improve QoS prediction performance in sparse graph settings.
Inspired by Liu et al. \cite{liu2023qosgnn} work, we adopt a graph attention mechanism in our model to better capture intricate user-service relationships in the QoS prediction context.

Unlike the approach used in QoSGNN\cite{liu2023qosgnn}, we did not directly incorporate an Attention module into the embedding matrix. Instead, we employed a graph attention mechanism, which performs better in processing graph-structured data. The graph attention mechanism, first proposed by Veličković et al. \cite{velivckovic2017graph}, applies self-attention to graph-structured data, allowing the model to learn complex node interactions adaptively. A key advantage of GAT lies in its ability to assign adaptive attention weights to each node's neighbors, thereby capturing fine-grained dependencies in the graph. This mechanism has enabled GAT to excel in tasks such as graph classification and node classification \cite{verma2023bet}. Various extensions of the graph attention mechanism have been proposed, such as the Heterogeneous Graph Attention Network (HGAT) \cite{HGAT} and Graph Attention Convolutional Networks \cite{liu2019deep}. Our work focuses on multi-order graph attention, as it enables broader contextual information propagation, which is particularly beneficial in extremely sparse matrices. We integrate this technique into our QoS prediction framework to evaluate its effectiveness in sparse data environments.


Although graph attention mechanisms significantly enhance the model's ability to capture structural information, their robustness remains limited under highly sparse and noisy conditions. To address this issue, we introduce an adversarial interaction module inspired by the concept of Generative Adversarial Networks (GANs) \cite{GANs}. Specifically, the generator produces QoS predictions based on latent embeddings, while the discriminator is trained to distinguish between real and synthetic interaction patterns, thereby encouraging the model to learn more realistic and noise-resistant representations. Furthermore, to handle the discrete nature of QoS interactions, which do not conform to the traditional Gaussian distribution, we incorporate the Gumbel-Softmax technique to generate differentiable pseudo-interaction samples. This mechanism facilitates adversarial learning in a discrete setting and enables the proposed QoSMGAA framework to achieve improved prediction robustness under sparse and noisy conditions.

%% file: text/3FQDQ.tex
\subsection{Question Definition}

In this section, we formally define the QoS prediction problem. Generally speaking, QoS prediction estimates service performance parameters for user-service interactions.

\begin{figure}[htbp]
    \centerline{\includegraphics[width = 0.8\textwidth]{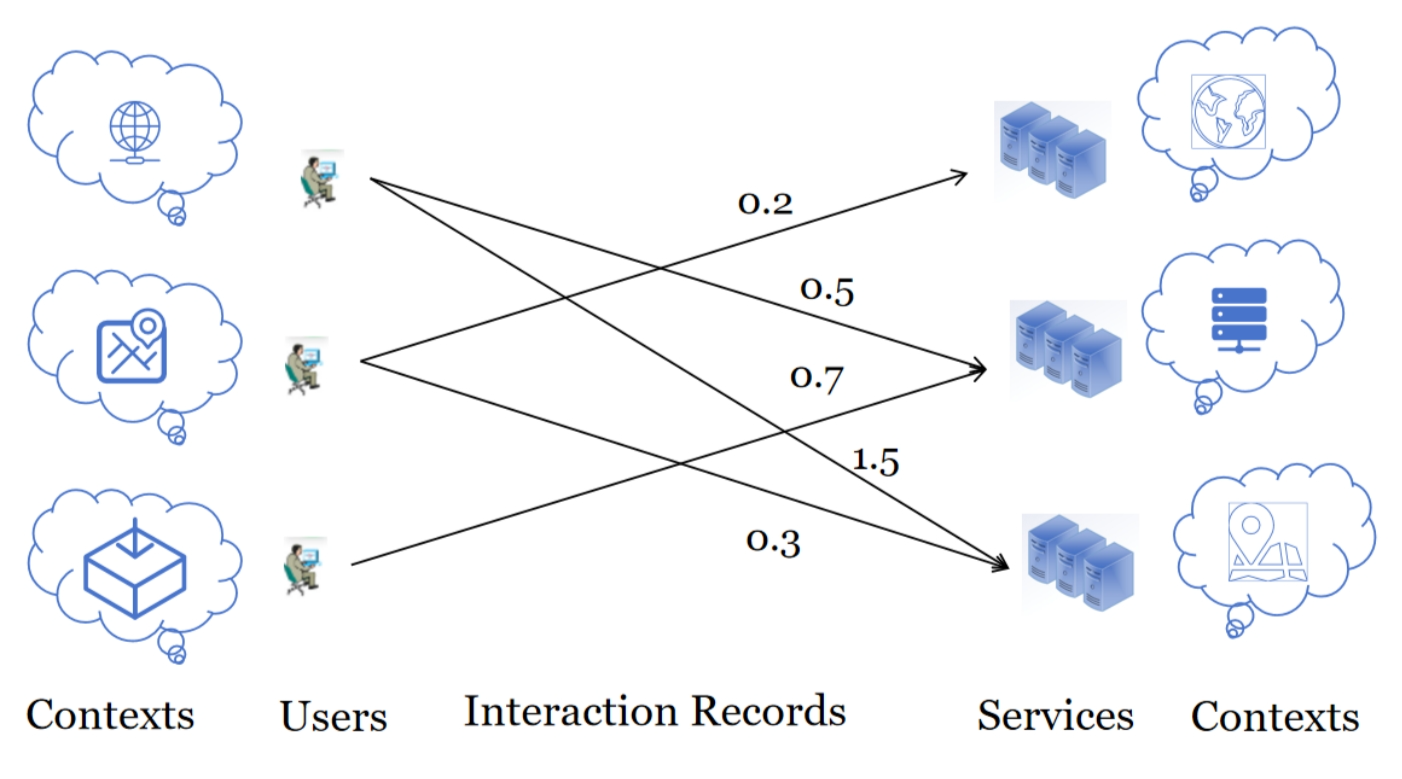}}
    \caption{An illustration of user-service interactions and associated QoS parameters}  
    \label{QoS} 
\end{figure}

Let $U=\left\{u_1, u_2, \ldots, u_m\right\}$ denote the set of users, and $S=\left\{s_1, s_2, \ldots, s_n\right\}$ denote the set of services. For each user $u \in U$ and service $s \in S$ is associated with a contextual feature set $C_{u, s}$ which includes auxiliary information. In addition, the known QoS record set $Q=\left\{q_{u, s}\right\}$, where $q_{u, s} \in \mathbb{R}^2$, represents the observed non-functional quality parameters when user $u$ interacts with service $s$. The objective of QoS prediction is to estimate the unknown values of $q_{u, s}$ for unobserved user–service pairs, potentially leveraging contextual features $C_{u, s}$ to improve prediction accuracy.
Formally, the prediction problem can be expressed as:

\begin{equation}
   q_{u,s} = f(U,S,C_{u,s},Q) 
\end{equation}
where $ f( \cdot ) $ is a model learned from the known records to predict QoS outcomes based on user-service pairs and their contextual information.

\subsection{Framework Overview}

\begin{figure}[htbp]
    \centering
    \includegraphics[width=\textwidth]{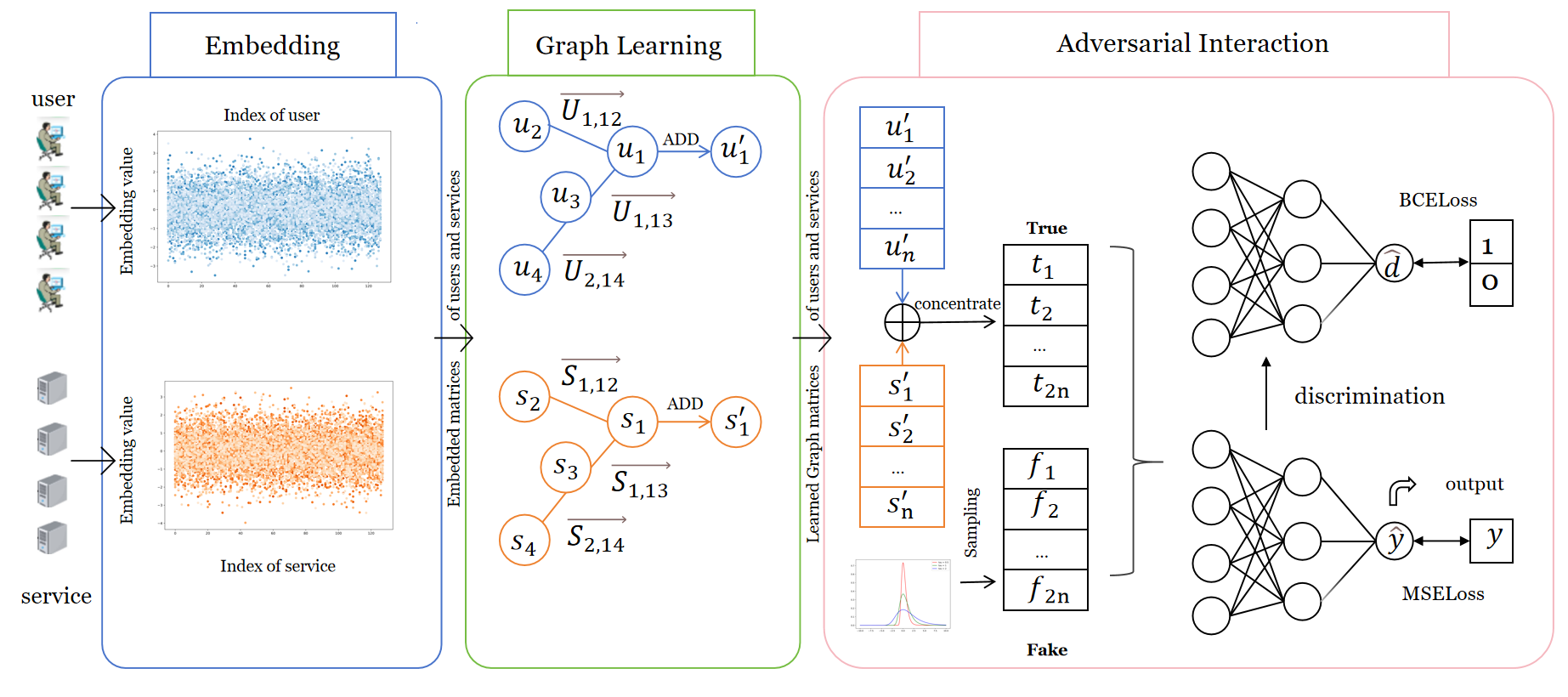}\vspace{-1em}
    \caption{Framework of QoSMGAA.} \vspace{-1em}
    \label{Fame}
\end{figure}

Figure \ref{Fame} demonstrates the structure of QoSMGAA, which can be delineated into three critical components: the embedding module, the graph learning module, and the interactive adversarial module. 

The Graph embedding module takes index values and retrieves the corresponding vectors from the factor matrices. To capture higher-order complex relationships in the graph, we designed a multi-order graph attention module in the second section to capture higher-order complex relationships. After processing the graph matrices with multi-order graph attention, they are put into the adversarial interaction module. This module leverages the principles of Generative Adversarial Networks, significantly enhancing the model's adaptability to new scenarios and prediction accuracy through continuously optimizing the generator and discriminator. We obtain our final prediction output after optimizing the graph attention and adversarial interaction through the training module.

%% file: text/4Solution.tex
In this section, we will elaborate on the key design elements of our model. 

\subsection{Sparse Sampling and Graph Construction}

The first part introduces how to sample a sparse matrix for our training and inference stages. $Q$ denotes the original user-service interaction matrix, where each element $q_{ij}$ represents an observed QoS value between user $i$ and service $j$. A non-zero entry $q_{ij} \neq 0$ indicates the presence of an actual interaction. The set $\Omega=\left\{(i, j) \mid q_{ij} \neq 0\right\}$ contains the indices of all such observed entries, and $|\Omega|$ denotes its cardinality, i.e., the total number of observed interactions.

Given a target sparsity level $\rho \in(0,1)$, we randomly permute $\Omega$ and retain only the first $\lfloor\rho|\Omega|\rfloor$ index pairs to form a new subset $\Omega_\rho$. This procedure ensures that the number of retained entries accounts for approximately a fraction $\rho$ of all observed interactions. Accordingly, $X_\rho$ represents the resulting sparse matrix, whose entries are defined by the rule.

\begin{equation}
    Q_\rho(i, j)= \begin{cases}q_{i j}, & \text { if }(i, j) \in \Omega_\rho, \\ 0, & \text { otherwise } .\end{cases}
\end{equation}
Where, if a pair $(i, j)$ is included in the retained index set $\Omega_\rho$, its original value $q_{i j}$ is preserved; otherwise, the corresponding position is set to zero. The resulting matrix $X_\rho$ thus maintains the original value distribution among the retained entries while achieving the desired global sparsity level $\rho$.

Moreover, we constructed heterogeneous graphs separately for users and services to characterize the rich attribute relationships between these entities comprehensively. The construction process of these graphs involves defining nodes and edges across multiple feature dimensions. Initially, unique identifiers were assigned to each user and service entity, facilitating clear differentiation within the graph structure. Subsequently, additional attribute nodes corresponding to distinct properties of users and services, such as geographical regions and service providers, were established. A globally unique mapping from feature values to node identifiers was created to ensure the uniqueness and integrity of each attribute representation within the graph.

Following the node creation, edges were constructed to capture explicit relationships between entities and their corresponding attribute nodes. Specifically, edges were formed between an entity node (user or service) and attribute nodes when the entity possessed specific characteristics. The graph edges were designed as bidirectional connections to capture semantic relationships fully, and self-loops were added to each node to enhance graph stability and strengthen node feature representation capabilities. Consequently, the resulting heterogeneous graphs encapsulate direct relationships between entities and their attributes, providing rich, structured information beneficial for subsequent graph neural network-based embedding learning.

\subsection{Embedding}

We use $U$ and $S$ to represent the user and service indexes. To improve the model’s performance and gain deeper insights into the complex dynamic relationships between nodes, we map user and service nodes to a high-dimensional feature space. Specifically, we transform these nodes into high-dimensional feature vectors through an embedding layer. The following formulas can describe this transformation:

\begin{equation}
    \label{eq5.1.1}
       U_e=\operatorname{Embedding}\left(W_u, U\right) ,
    \end{equation}
\begin{equation}
    \label{eq5.1.2}
        S_e=\operatorname{Embedding}\left(W_s, S\right) ,
    \end{equation}
where $W_u$ and $W_s$ are the embedding matrices for users and services, respectively. The dimensions of these matrices are defined by the number of nodes in each respective matrix and the embedding dimension $R$. $U_e$ and $S_e$ represent the embedded user and service feature vectors, where $U_e \in \mathbb{R}^{U_n \times R}$ and $S_e \in \mathbb{R}^{S_n \times R}$. $U_n$ and $S_n$ refer to the number of user and service nodes in the graph, respectively.

\subsection{Graph Learning}

To enable effective learning from the user and service graphs $U_e$ and $S_e$, we introduce a multi-order graph attention mechanism that captures dynamic and long-range dependencies among nodes. Unlike traditional graph attention networks that rely primarily on immediate neighbors, our approach aggregates information from multi-hop neighborhoods, thereby enhancing the model’s representational capacity under sparse and noisy conditions. As outlined in Algorithm \ref{alg:simplified-user-graph} and Equations \ref{eq5.1.8} and \ref{eq5.1.9}, attention coefficients are computed for each node’s n-hop neighbors and used to perform weighted aggregations, integrating richer contextual information.


\begin{algorithm}
\caption{Simplified Graph Learning for Users}
\label{alg:simplified-user-graph}
\begin{algorithmic}[1]
\State \textbf{Input:} List of user IDs $u_i \in \text{Users } \mathcal{U}$
\State \textbf{Output:} Updated user feature matrix $U' = \{ u_i' \}$
\For{each neighbor node $u_j \in \mathcal{N}_i^d$ across all orders $d$}
    \State $U_{d,ij} \gets \textsc{MultiOrderGAT}(u_i, u_j)$ \Comment{Compute attention by Eq.~\ref{eq5.1.8}}
    \State $u_i' \gets u_i' + \textsc{Aggregate}(u_j, U_{d,ij})$
\EndFor
\State \Return $U'$
\end{algorithmic}
\end{algorithm}

For each node $ i $, we first compute the similarity $ e_{ij}$ between node $i$ and its neighbor node $ j $ using the following formula:

\begin{equation}
    \label{eq5.1.3}
        e_{i j}=a\left(\left[{W}_h h_i \| {W}_h h_j\right]\right), \quad j \in \mathcal{N}_i ,
    \end{equation}
where $h_i$ and $h_j$ denote the feature vectors of nodes $i$ and $j$, respectively, $|$ denotes vector concatenation, and $\mathcal{N}_i$ is the set of neighbor nodes of node $i$. The function $a$ maps the concatenated high-dimensional features to a real number. The weight parameter ${W}_h$ involved in this mapping is learnable, ensuring the model can adaptively optimize feature representations.

Subsequently, we use the softmax normalization operation to normalize the similarity $e_{ij}$ and compute the attention weight $\alpha_{ij}$ of neighbor nodes for node $i$:

\begin{equation}
    \label{eq5.1.4}
        \alpha_{i j}=\frac{\exp \left(\operatorname{LeakyReLU}\left(e_{i j}\right)\right)}{\sum_{k \in \mathcal{N}_i} \exp \left(\operatorname{LeakyReLU}\left(e_{i k}\right)\right)} ,
    \end{equation}
Where the activation function LeakyReLU is defined as:

\begin{equation}
    \label{eq5.1.5}
       \operatorname{LeakyReLU}(x)=\max (0, x)+\mathbf{ leak } \cdot \min (0, x) ,
    \end{equation}
where the $\mathbf{ leak }$ is a hyperparameter. LeakyReLU helps maintain a non-zero gradient for the information flow and prevents the gradient from vanishing during training.

Expanding the above formulas to our graph structures $U$ and $S$, we obtain the attention coefficients $U_{ij}$ and $S_{ij}$ between users and services as follows:

\begin{equation}
    \label{eq5.1.6}
U_{i j}=\frac{\exp \left(\operatorname{Leaky} \operatorname{ReLU}\left(\tilde{{a}}^T\left[{W}_{{u}} {u}_{{i}} \| {W}_{{u}} {u}_{{j}}\right]\right)\right)}{\sum_{k \in \mathcal{N}_i} \exp \left(\operatorname{LeakyReLU}\left(\tilde{{a}}^T\left[{W}_{{u}} {u}_{{i}} \| {W}_{{u}} {u}_{{k}}\right]\right)\right)} ,
    \end{equation}


\begin{equation}
    \label{eq5.1.7}
S_{i j}=\frac{\exp \left(\operatorname{LeakyReLU}\left(\tilde{{b}}^T\left[{W}_{{s}} {s}_{{i}} \| {W}_{{s}} {s}_{{j}}\right]\right)\right)}{\sum_{k \in \mathcal{N}_i} \exp \left(\operatorname{LeakyReLU}\left(\tilde{{b}}^T\left[{W}_{{s}} {s}_{{i}} \| {W}_{{s}} {s}_{{k}}\right]\right)\right)} ,
    \end{equation}
where ${u}_i$, ${u}_j$, and ${s}_i$, ${s}_j$ denote the vector representations of users and services in the embedding space. ${W}{{u}}$ and ${W}{{s}}$ are weight matrices for the feature transformation of users and services, used to adjust each node's feature vector in the attention mechanism.
$\tilde{{a}}$ and $\tilde{{b}}$ are used to project node features onto a real number through a linear transformation to compute the unnormalized attention coefficients. The LeakyReLU activation function introduces non-linearity, increases the model's expressive power, and helps mitigate gradient vanishing issues.

The multi-order graph attention mechanism aims to capture the complex interactions between users and services through multi-layer signal propagation. Specifically, we extend the attention mechanism to multiple graph layers, allowing the model to capture and integrate node information at different layers.
More specially, We define $d$-order attention coefficients $U_{d,ij}$ and $S_{d,ij}$ as follows:

\begin{equation}
    \label{eq5.1.8}
U_{d, i j}=\frac{\exp \left(\operatorname{Leaky} \operatorname{ReLU}\left(\tilde{{a}}^T\left[{W}_{{u}} {u}_{{i}} \| {W}_{{u}} {u}_{{j}}\right]\right)\right)}{\sum_{k \in \mathcal{N}_i^d} \exp \left(\operatorname{LeakyReLU}\left(\tilde{{a}}^T\left[{W}_{{u}} {u}_{{i}} \| {W}_{{u}} {u}_{{k}}\right]\right)\right)} ,
    \end{equation}

\begin{equation}
    \label{eq5.1.9}
S_{d,i j}=\frac{\exp \left(\operatorname{LeakyReLU}\left(\tilde{{b}}^{\top}\left[{W}_{{s}} {s}_{{i}} \| {W}_{{s}} {s}_{{j}}\right]\right)\right)}{\sum_{k \in \mathcal{N}_i^d} \exp \left(\operatorname{LeakyReLU}\left(\tilde{{b}}^{\top}\left[{W}_{{s}} {s}_{{i}} \| {W}_{{s}} {s}_{{k}}\right]\right)\right)} ,
    \end{equation}
where $\mathcal{N}_i^d$ denotes the set of neighbor nodes of node $i$ in the $d$-order graph structure. This way, our model can capture and integrate information into the graph.

After computing the attention weights, we obtain the new feature vector for each node by performing a weighted sum of its feature vector and the feature vectors of its neighboring nodes. Formally, the update formulas for the feature vectors can be written as:

\begin{equation}
    \label{eq5.1.10}
{u}_{{i}}^{\prime}=\sigma\left(\sum_{d=1}^d \sum_{k \in N_i^d} U_{d,ij}W_{u} u_k\right) ,
    \end{equation}

\begin{equation}
    \label{eq5.1.11}
{s}_{{j}}^{\prime}=\sigma\left(\sum_{d=1}^d \sum_{k \in N_i^d} U_{d,ij}W_{s} s_k\right) ,
    \end{equation}
where ${u}_{{i}}^{\prime}$ and ${s}_{{j}}^{\prime}$ represent embedded user or service matrix that has finished the message passing stage and $d$ is the number of Order of Multi-Order Grapgh Attention.


    
To help the model better focus on the important parts of the input user embeddings and peer embeddings, similar to Vaswani et al.\cite{vaswani2017attention}, we now extend our mechanism to use multi-head attention. This mechanism learns different input representations by running multiple independent attention heads in parallel and then combines these representations to form the final feature vector for each node. This strategy enhances the model's ability to capture information.

\begin{equation}
    \label{eq5.1.14}
{u}_{{i}}^{\prime}= \|_{n=1}^N \sigma\left(\sum_{d=1}^d \sum_{k \in N_i^d} U_{d,ij}^d W_{u}^d u_k\right) ,
\end{equation}

\begin{equation}
    \label{eq5.1.15}
{s}_{{j}}^{\prime}= \|_{n=1}^N \sigma\left(\sum_{d=1}^d \sum_{k \in N_j^d} U_{d,ij}^d W_{s} ^d u_k\right) ,
    \end{equation}
where $N$ is the number of attention heads.

\subsection{Adversarial Interaction}

In our matrix interaction module, we focus on modeling the interaction between the user and service embedding vectors. Specifically, we use fully connected layers with adversarial neural networks to predict QoS missing values. This technique more effectively reveals complex interactions between users and services and significantly improves prediction accuracy in complex scenarios.

The procedure for the adversarial neural network is detailed in Algorithm \ref{alg:adv-network}. Real data is constructed by concatenating the learned embedding matrix with the service matrix. In contrast, fake data is generated by sampling from the Gumbel-Softmax distribution at varying densities. Lines 7-9 represent predictions made for real samples, followed by lines 12-14, where the prediction results are fed into the discriminator. A similar procedure is performed for the fake samples. The predicted values for real samples serve as the true output of the model, while the discriminator's output helps optimize both the predictor and the discriminator. This process endows our model with an enhanced ability to capture high-order relationships between users and services while also mitigating the impact of noise.



Specifically, after the graph learning process, we obtain the updated user graph matrix and service graph matrix ${u}_{{i}}^{\prime} \in {U_e}^{\prime}$ and ${s}_{{j}}^{\prime} \in {S_e}^{\prime}$. To effectively capture the complex high-order relationships in these matrices, we first concatenate these two matrices to form a unified interaction matrix ${T}$ as follows:

\begin{equation}
    \label{eq5.1.16}
{T}={U_e}^{\prime} \| {S_e}^{\prime},
    \end{equation}
where ${U_e}^{\prime}$ and ${S_e}^{\prime}$ represent the user and service matrices after graph attention. The matrix ${T}$ serves as the actual input embedding for the prediction layer, providing a method to integrate user and service features. Based on this, we generate a fake input embedding.

When handling discrete data, the learning efficiency of the model is often limited by the differentiability of operations. The Gumbel Softmax technique overcomes this challenge, allowing for approximate sampling of discrete variables while maintaining end-to-end differentiability. Gumbel Softmax combines the Gumbel distribution and the Softmax function to generate approximately discrete outputs.

\begin{algorithm}
\caption{Adversarial Neural Network for User–Service Interaction}
\label{alg:adv-network}
\begin{algorithmic}[1]
\State \textbf{Input:} User graph matrix ${U_e}'$, Service graph matrix ${S_e}'$
\State \textbf{Output:} True predictions $\hat{y}_t$, False predictions $\hat{y}_f$, Discriminator outputs $\hat{d}_t$, $\hat{d}_f$

\State Concatenate ${U_e}'$ and ${S_e}'$ to form interaction matrix $T$
\State Generate fake input $F$ using Gumbel-Softmax

\Comment{Prediction Layer for True Input}
\State $y_{h1} \gets \textsc{ReLU}(\textsc{LayerNorm}(\textsc{MLP}(T)))$
\State $y_{h2} \gets \textsc{ReLU}(\textsc{LayerNorm}(\textsc{MLP}(y_{h1})))$
\State $\hat{y}_t \gets \textsc{MLP}(y_{h2})$

\Comment{Prediction Layer for Fake Input}
\State $y_{f1} \gets \textsc{ReLU}(\textsc{LayerNorm}(\textsc{MLP}(F)))$
\State $y_{f2} \gets \textsc{ReLU}(\textsc{LayerNorm}(\textsc{MLP}(y_{f1})))$
\State $\hat{y}_f \gets \textsc{MLP}(y_{f2})$

\Comment{Discriminator Layer for True Input}
\State $d_{h1} \gets \textsc{LeakyReLU}(\textsc{MLP}(\hat{y}_t))$
\State $d_{h2} \gets \textsc{BatchNorm}(\textsc{LeakyReLU}(\textsc{MLP}(d_{h1})))$
\State $\hat{d}_t \gets \textsc{MLP}(d_{h2})$

\Comment{Discriminator Layer for Fake Input}
\State $d_{h1} \gets \textsc{BatchNorm}(\textsc{LeakyReLU}(\textsc{MLP}(\hat{y}_f)))$
\State $d_{h2} \gets \textsc{BatchNorm}(\textsc{LeakyReLU}(\textsc{MLP}(d_{h1})))$
\State $\hat{d}_f \gets \textsc{MLP}(d_{h2})$

\State \Return $\hat{y}_t$, $\hat{y}_f$, $\hat{d}_t$, $\hat{d}_f$
\end{algorithmic}
\end{algorithm}


The traditional Softmax function converts a real-valued vector into a probability distribution: ${F}$ with the same dimension as ${T}$ for the network training process to apply adversarial learning strategies.






\begin{equation}
    \label{eq:gumbel_input}
    Z \sim \mathcal{N}(0, 1) \in \mathbb{R}^{B \times 2d}
\end{equation}

We initialize $Z$ as a random matrix drawn from a standard Gaussian distribution, where $B$ is the batch size and $2d$ corresponds to the combined dimensionality of user and service embeddings.

\begin{equation}
    \label{eq:gumbel_noise}
    G = -\log\left(-\log(U)\right), \quad U \sim \mathcal{U}(0, 1) \in \mathbb{R}^{B \times 2d}
\end{equation}

To simulate sampling from a discrete distribution in a differentiable manner, we add Gumbel noise $G$ to $Z$ and apply the softmax function with temperature $\tau$:

\begin{equation}
    \label{eq:gumbel_sample}
    F = \operatorname{Softmax}\left((Z + G)/\tau\right)
\end{equation}
where, $F \in \mathbb{R}^{B \times 2d}$ represents the resulting pseudo-sample generated via the Gumbel-Softmax mechanism. The temperature parameter $\tau$ controls the smoothness of the output distribution.

We constructed a neural network prediction layer based on a Multi-Layer Perceptron (MLP). This structure enhances the model's nonlinear learning capability by processing information in layers using Layer Normalization (LayerNorm)\cite{ba2016layer} and ReLU activation functions. The computation process is as follows:

\begin{equation}
    \label{eq5.1.22}
y_{h1} = ReLu(LayerNorm(MLP(y))) ,
    \end{equation}
\begin{equation}
    \label{eq5.1.23}
y_{h2} = ReLu(LayerNorm(MLP(y_{h1}))) ,
    \end{equation}
\begin{equation}
    \label{eq5.1.24}
\hat{y} = MLP(y_{h2}) ,
    \end{equation}
where $y_{h1}$ and $y_{h2}$ are the hidden layer of the fully connected layer, we define the input of $y$ as determined by two main tensors, ${F}$ (false input) and ${T}$ (true input), which generate the corresponding true prediction $\hat{y}_t$ and false prediction $\hat{y}_f$. 
This configuration allows us to optimize the parameters of the prediction layer formulas (\ref{eq5.1.22}- \ref{eq5.1.24}) in an adversarial framework by comparing the true and false predictions.

To further improve the prediction accuracy, the true prediction $\hat{y}_t$ and the false prediction $\hat{y}_f$ are input into the discriminator. The discriminator employs BatchNorm\cite{ioffe2015batch} techniques to enhance the model's generalization ability and uses LeakyReLU for nonlinear activation, implemented as follows:

\begin{equation}
    \label{eq5.1.25}
d_{h1} = BatchNorm(LeakyReLU(MLP(d))) ,
    \end{equation}
\begin{equation}
    \label{eq5.1.26}
d_{h2} = BatchNorm(LeakyReLU(MLP(d_{h1}))) ,
    \end{equation}
\begin{equation}
    \label{eq5.1.27}
\hat{d} = MLP(d_{h2}) ,
    \end{equation}
where the input $d$ is composed of $\hat{y}_f$ and $\hat{y}_t$, and the resulting outputs $\hat{d}_f$ and $\hat{d}_t$ correspond to the discrimination results for the false and true predictions, respectively. $d_{h1} $ and$ d_{h2}$ refer to the hidden layer of Discrimination model. In this way, the discriminator's loss function is used to optimize both the prediction layer and the discriminator itself, enhancing the effectiveness of adversarial training.

The main function of BatchNorm is to normalize each batch of data so that its mean is zero and variance is one. This process can reduce the impact of input distribution changes between different layers (i.e., internal), accelerate the model's training process, and improve convergence speed.

\subsection{Model Training}
After detailing the various modules, this section discusses how to train our model to achieve full matrix completion. Our core training objective is to minimize the loss function $L$ to optimize the overall performance of matrix completion.

The overall loss function of the model is defined as follows:
\begin{equation}
    \label{eq5.1.28}
L = \lambda f(\hat{d}_t , y_t )  + (1-\lambda ){g(\hat{y}_t ,y)} ,
    \end{equation}
where the function $f$ represents the binary cross-entropy loss (BCEWithLogitsLoss), and $g$ represents the mean square error loss (MSELoss). In this case, $\hat{y}_t$ is the predicted value by the model, $\hat{d}_t$ is the discriminator's prediction of the true sample, $y_t$ is the actual label, and $y$ is the true value from the dataset. $\lambda$ is a hyperparameter that balances the cross-entropy loss and mean square error loss. Meanwhile, the loss function of the discriminator is defined by the following equation:

\begin{equation}
    \label{eq5.1.29}
L_d =  {f(\hat{d}_t , y_t ) } + {f(\hat{d}_f ,y_f)} ,
    \end{equation}
where $\hat{d}_f$ is the discriminator's prediction output for the false sample, and $y_f$ is the label for the false sample. The overall training process is carried out in an alternating optimization manner, with the discriminator and generator updated at a frequency of 1:1 until the overall $L$ converges. The loss functions ${f}$ and ${g}$ are defined as follows:

\begin{equation} 
\label{eq5.1.30}
f(x, y) = - \left[ y \log(\sigma(x)) + (1 - y) \log(1 - \sigma(x)) \right],
\end{equation}
where $\sigma(x) = 1 /(1 + \exp (-x)) , $

\begin{equation}
    \label{eq5.1.31}
g(x,y) = (x-y)^2
   \end{equation}




%% file: text/5Exp.tex
This section provides a comprehensive overview of the experimental design and the performance evaluation criteria for validating the proposed model framework. Our research aims to explore the following key questions:

\begin{itemize}
    \item \textbf{RQ1} How does our model perform compared to other baseline models on real-world datasets? 
    \item \textbf{RQ2} How does the multi-order attention mechanism affect the model's performance (Ablation Study)?
    \item \textbf{RQ3} What is the impact of the adversarial neural network on the model's effectiveness (Ablation Study)?
    \item \textbf{RQ4} How do different negative sampling strategies affect the model's performance (Ablation Study)?
    \item \textbf{RQ5} How do hyperparameters influence our model?
    \item \textbf{RQ6} What are the proposed model's runtime efficiency and memory consumption characteristics during inference?
    \item \textbf{RQ7} How well can our QoSMGAA handle noisy and sparse data compared to baseline methods?
\end{itemize}

\subsection{Datasets}

We utilized the widely adopted public WsDream dataset\cite{zheng2012investigating}, which serves as a standard benchmark for evaluating QoS prediction models. The dataset comprises two main parts: users' response time (RT) data for Web services and throughput (TP) data. Overall, the WsDream dataset includes invocation records from users worldwide. The dataset comprises invocation records collected globally, where users are distributed across multiple geographic regions and services are hosted on diverse cloud-based server environments. It contains detailed logs of user-service interactions, including measured response times and throughput.

To further evaluate our model's generalization capability, we adopted the EEL \cite{10660479}  dataset. It comprises 5,174 end-to-end latency measurement points collected from November 27 to December 17, 2021. In total, it shall consist of 94,319,155 interaction records. For our experiments, we selected two key non-functional attributes from this dataset:(DELAY) tcp\_out\_delay, representing the Round-Trip Time (RTT) of a probe packet, and hops(HOPS), indicating the number of intermediate nodes between the source and destination machines. Additionally, contextual information such as the geographical regions and service providers of both the source and target servers was included in the datasets.

To investigate the influence of data sparsity on the prediction performance, we adopted different training density settings across two benchmark datasets. Specifically, for the WSDream dataset, four training densities were considered: 2.5\%, 5\%, 7.5\%, and 10\%. In contrast, due to the larger scale and redundancy of the EEL dataset, we selected three levels: 2.5\%, 5\%, and 7.5\%. For each selected density level, the remaining data samples were used as the test set to evaluate the model's performance at different training densities.

\subsection{Metrics}
We adopted two commonly used evaluation metrics to assess model performance in QoS prediction.
\begin{itemize}
    \item {Mean Absolute Error (MAE):

\begin{equation}
    \label{eq6.1}
\mathrm{MAE}=\frac{1}{N} \sum_{ij=1}^N\left|y_{ij}-\hat{y}_{ij}\right|
\end{equation}

where $y_{ij}$ represents the $ij$-th observed value, $\hat{y}_{ij}$ represents the $ij$-th predicted value, and $N$ is the total number of samples. MAE measures the average absolute difference between the predicted and actual values. A smaller MAE indicates higher prediction accuracy and lower error.}
\item {
Root Mean Square Error (RMSE):

\begin{equation}
    \label{eq6.2}
\mathrm{RMSE}=\sqrt{\frac{1}{N} \sum_{ij=1}^N\left(y_{ij}-\hat{y}_{ij}\right)^2}
\end{equation}

RMSE is the square root of the mean of the squared prediction errors, assigning higher penalty weights to larger errors. Therefore, RMSE is a more sensitive metric to significant errors, making it especially suitable for applications where considerable prediction deviations are critical.
}
\end{itemize}

\subsection{Baselines}

To verify the effectiveness and superiority of the proposed model, we selected ten representative baseline methods, including three CF models, two MF models, three deep learning (DL) models, and two GNN models:

\begin{itemize}
\item \textbf{UPCC} \cite{UPCC}: CF-based method predicting missing values using QoS of the top $k$ similar users identified via Pearson correlation.
\item \textbf{IPCC} \cite{IPCC}: CF-based method predicting QoS based on the top $k$ similar services.
\item \textbf{UIPCC} \cite{UIPCC}: CF-based method that integrates user and service similarities (UPCC and IPCC) to enhance QoS prediction accuracy.
\item \textbf{PMF} \cite{mnih2007probabilistic}: MF-based probabilistic model to uncover latent user-service features for QoS prediction.
\item \textbf{BiasMF} \cite{yu2014personalized}: MF-based model incorporating bias terms to capture user, service, and contextual information.
\item \textbf{CSMF} \cite{wu2018collaborative}: DL-based model employing context-aware embedding learning to extract features for QoS prediction.
\item \textbf{NFMF} \cite{xu2021nfmf}: DL-based model using fully-connected networks and multi-task learning for QoS prediction.
\item \textbf{NCRL} \cite{zou2022ncrl}: DL-based dual-tower residual network integrating multi-layer perceptron (MLP) to predict QoS.
\item \textbf{GraphMF} \cite{li2020graphmf}: GNN-based model employing graph convolutional networks (GCN) to capture latent user-service interactions.
\item \textbf{QoSGNN} \cite{liu2023qosgnn}: GNN-based model combining attention mechanisms with graph neural networks to dynamically capture user-service interactions.
\end{itemize}

\subsection{Experimental Setup}

Our dataset was divided into training, testing, and validation sets. We conducted three rounds of experiments for each result and averaged the final results across the three rounds. In each round, we set 100 epochs. Each epoch involved training on the training set and evaluation on the validation set. The validation results were recorded as the performance for that epoch. 
The model parameters that achieved the best validation performance during training were selected to evaluate the final test results for this round.


In our model, the order of graph attention is set to 2, and the matrix's embedding dimension is fixed at 32 throughout the process, from initial embedding to graph learning. The adversarial interaction module's hidden layer dimension was set to 128. In the discriminator, we use a hidden dimension of 4 for autoregressive learning. 

To mitigate overfitting, we employed an early stopping strategy with a patience threshold of 15 epochs based on validation loss. And, we incorporated a stochastic dropout\cite{srivastava2014dropout} mechanism in the multi-order graph attention learning module, with a dropout rate of 0.1, randomly omitting 10\% of the feature activations during training. The batch size was set to 128, and the AdamW\cite{loshchilov2017decoupled} optimizer was employed for parameter updates during model training.

\subsection{Comparison Experiments(RQ1)}



\input{text/rt_new_table}
\input{text/tp_new_table}
\input{text/de_new_table}
\input{text/ho_new_table}

\begin{itemize}

    \item Collaborative Filtering-based methods (UPCC, IPCC, UIPCC) perform poorly under sparse data because they rely on dense user-service interaction data to calculate similarities. Height data sparsity leads to inaccurate similarity calculations, negatively impacting prediction performance.

    \item Matrix Factorization-based methods (PMF, biasMF) underperform in practice due to their reliance on linear assumptions, which overlook the non-linear complexities of user behavior and service quality. Their effectiveness is further limited in extremely sparse data scenarios.

    \item Deep learning-based models, such as Neural Factorization Machine (NFMF) and Neural Collaborative Regression (NCRL), struggle to capture the complex user-service relationships, relying too much on shallow interactions and linear factors, which limits their effectiveness with graph-structured data.

    \item GNNs-based methods, such as GraphMF and QoSGNN, utilize graph neural networks and collaborative filtering to accurately predict missing QoS values, especially effective in low-density matrices. However, they rely on high-quality graph inputs and may falter with noisy or unfamiliar graph structures.

    \item QoSMGAA integrates multi-order graph attention mechanisms and adversarial networks, showing outstanding improvements in model robustness and prediction accuracy. This is attributed to our superior model architecture, where multi-order graph attention mechanisms effectively propagate sparse information. Additionally, the use of fully connected and adversarial neural networks during the interaction phase greatly improves the handling of interactions. Supported by these two robust architectures, our model demonstrated superior performance in comparative experiments.
\end{itemize}

In the comparison of the RT and TP datasets, QoSMGAA outperforms other models by at least 10\% in predicting missing QoS values in MAE, particularly demonstrating its superiority in high sparsity.

In the RT dataset, QoSMGAA shows significant performance gains over the following best method (QoSGNN). For instance, at a 2.5\% density, it improves MAE by 16.33\% and RMSE by 7.84\%. These improvements are consistent across all densities, and they can deal with scenarios with greater data sparsity.
In the TP dataset, QoSMGAA demonstrates effectiveness and scalability in sparse conditions. For example, at a 2.5\% density, it shows an improvement of 19.29\% in MAE and 5.98\% in RMSE over other methods, reinforcing its capability to efficiently handle different data volumes and structures.

Tables \ref{tab:delay_comparison_new} and \ref{tab:hops_comparison_newer} summarize the performance evaluation of various methods on non-functional QoS attributes, namely delay time and the number of hops, across different matrix density levels (2.5\%, 5\% and 7.5\%). For each setting, the mean and standard deviation of MAE and RMSE are calculated over three independent runs using different random seeds. Based on the reported results, several noteworthy insights can be drawn as follows:

\begin{itemize}
    \item In comparison to traditional approaches such as PMF, deep learning-based methods consistently exhibit superior performance across the two aforementioned datasets, particularly under conditions of data sparsity. Deep learning methods serve as an effective paradigm for capturing latent structural relationships between users and services within QoS prediction tasks. This demonstrates that deep learning techniques not only significantly enhance the prediction accuracy for missing QoS values but also effectively alleviate the cold-start problem inherent in QoS prediction scenarios.
    \item GNN-based methods leverage the construction of user-service bipartite graphs to more effectively aggregate the latent representations from neighboring interactions, thus providing an enhanced embedding learning strategy for individual nodes. Notably, our findings indicate that GNN-based approaches substantially improve predictions of non-functional attributes in user-service interactions compared to standard deep learning methods, particularly in sparse scenarios, thereby effectively mitigating the cold-start problem. The primary reason for this improvement lies in the availability of tons of contextual nodes within the current datasets, contributing to sufficient representation of users and services based on the graph structure, consequently enhancing the propagation of higher-order relational messages.
    \item To address the limitations of traditional graph neural networks in capturing higher-order relationships within graph structures, we propose the QoSMGAA framework, which employs a multi-order graph attention mechanism. This enables users and services to learn embeddings from higher-order graph structures, significantly enhancing feature vector learning capabilities. In the interaction modeling phase, the proposed adversarial network outperforms conventional CF methods by effectively filtering out noisy data and concurrently improving the robustness of user-service interaction modeling. Empirical evaluations on both datasets demonstrate that our QoSMGAA framework consistently achieves superior prediction performance.
\end{itemize}

On both the DELAY and HOPS datasets, our proposed framework consistently demonstrates strong predictive performance across all evaluated conditions. Specifically, on the DELAY dataset, the model achieves approximately a 5\%  improvement in MAE compared to the next-best baseline, QoSGNN. On the HOPS dataset, the observed performance gain exceeds 10\%, further highlighting the effectiveness of our architecture. Notably, under extremely sparse conditions, such as a matrix density of 2.5\%, our framework exhibits substantially larger improvements relative to other density settings, indicating its strong capability in handling highly incomplete QoS data. These results suggest that the proposed framework is well-suited for accurate QoS prediction in sparse environments.

In conclusion, although deep learning-based models generally outperform traditional deep learning models, like NCRL, they do not always perform better than conventional methods in practical applications, because it doesn't fully exploit the graph structure. This highlights the importance of selecting appropriate deep-learning architectures when handling complex graph-structured data. Graph deep learning models show great potential in QoS prediction, and our architecture benefits significantly from its expressiveness, producing better prediction results for missing values.

\subsection{Ablation Experinment}

To investigate how various components contribute to the performance of our model, we designed a series of ablation experiments targeting three aspects: (1) the multi-order GAT mechanism, (2) the adversarial interaction module, and (3) the negative sampling strategy based on discrete samples. All experiments were conducted to evaluate the model's prediction performance regarding MAE and RMSE across four datasets, namely RT, TP, DELAY, and HOPS, under different matrix density settings. Specifically, matrix densities of 2.5\%, 5\%, 7.5\%, and 10\% were used for RT and TP, whereas only 2.5\%, 5\%, and 7.5\% applied to DELAY and HOPS due to data availability limitations.

\subsubsection{Impact of Multi-order Attention Mechanism (RQ2)}

To assess the effectiveness of the proposed multi-order graph attention (MOGAT) mechanism, we compared it against models employing only a standard first-order(GAT) or a two-order attention mechanism.

\begin{figure}[htbp]
    \begin{minipage}[t]{0.49\linewidth}
        \centering
        \includegraphics[width=\textwidth]{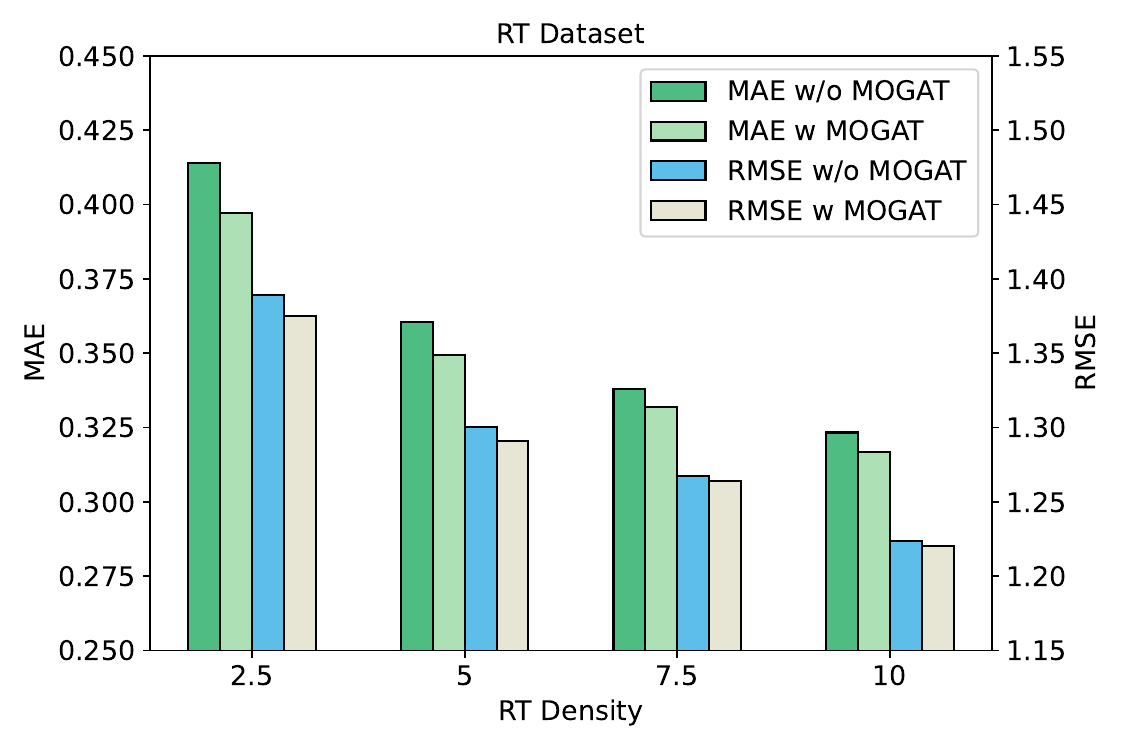}
        \centering 
    \end{minipage}
    \begin{minipage}[t]{0.49\linewidth}
        \centering
        \includegraphics[width=\textwidth]{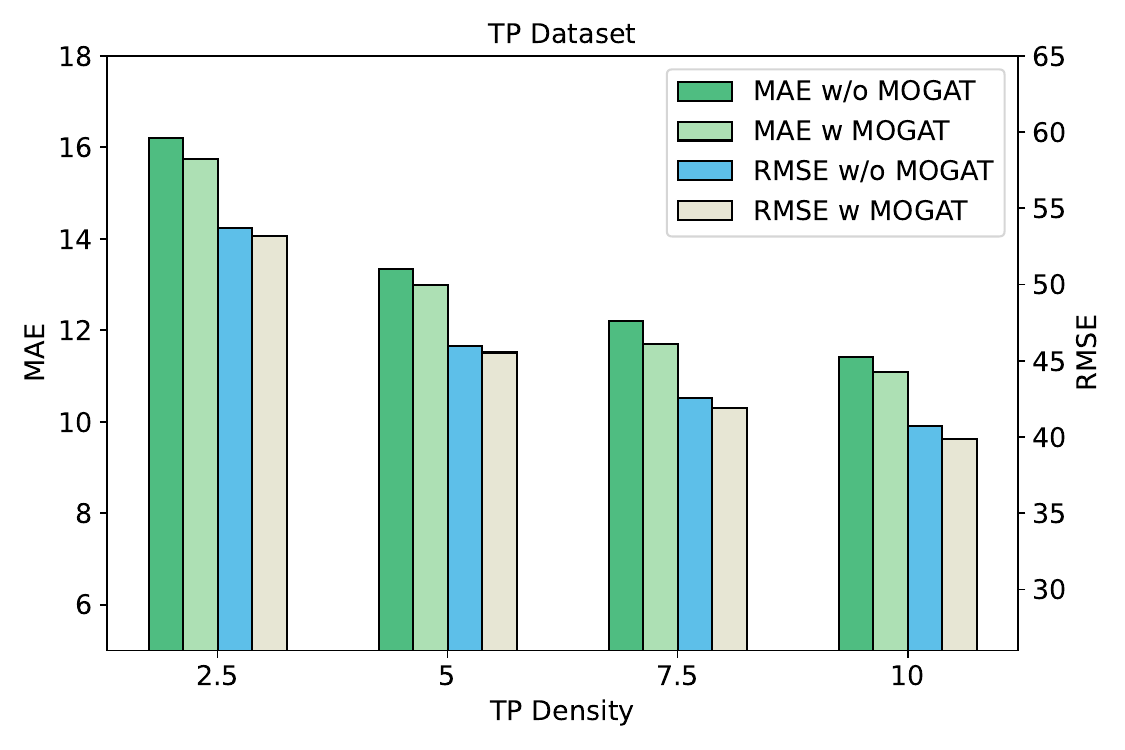}
        \centering  
    \end{minipage}
    \\
    \begin{minipage}[t]{0.49\linewidth}
    \centering
    \includegraphics[width=\textwidth]{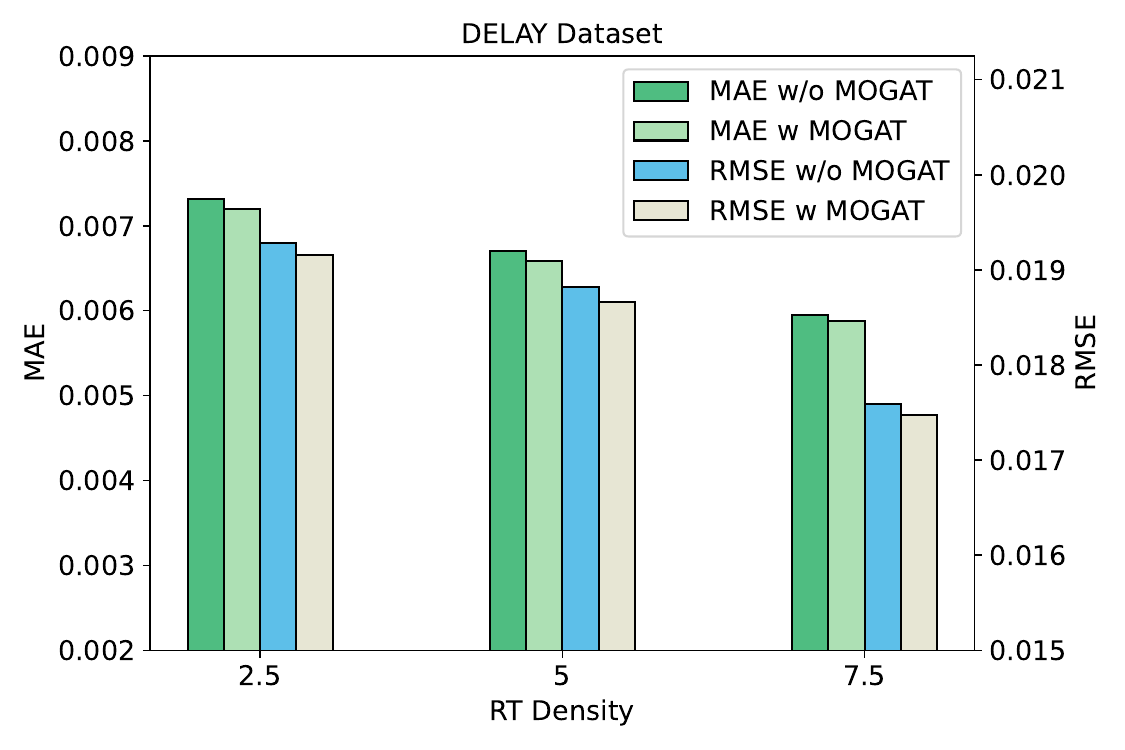}
    \centering   
\end{minipage}
\begin{minipage}[t]{0.49\linewidth}
    \centering
    \includegraphics[width=\textwidth]{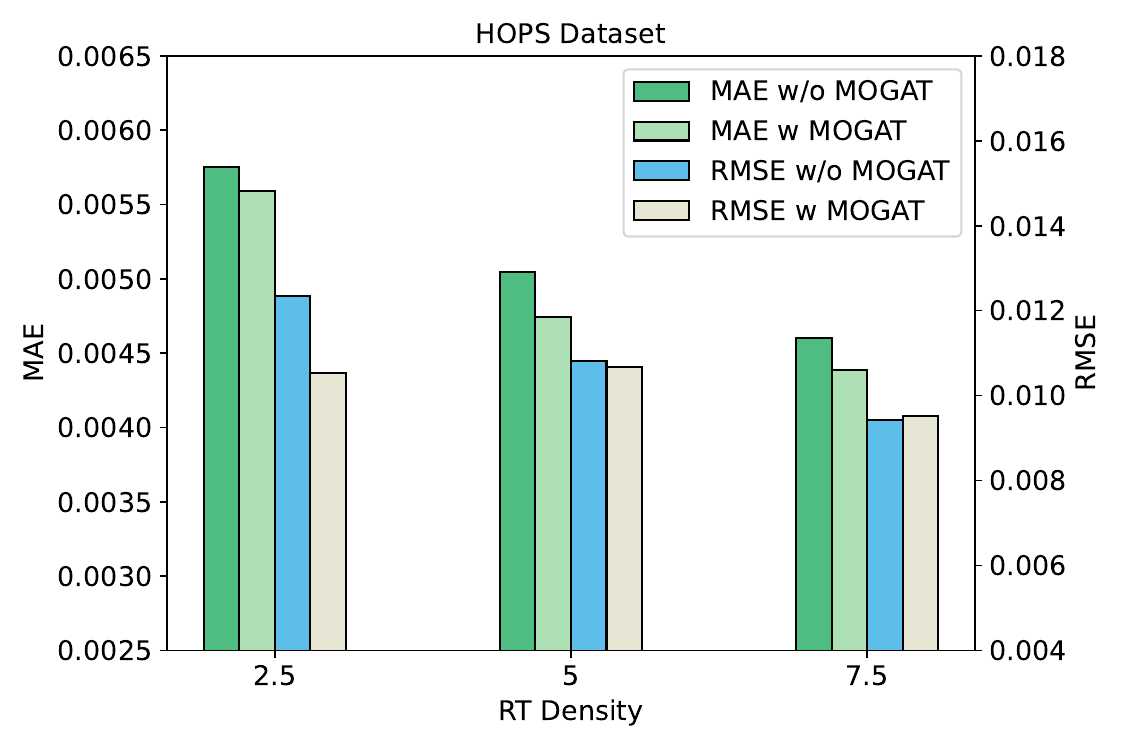}
    \centering   
\end{minipage}

    \caption{Ablation study on MOGAT}
    \label{fig1}
\end{figure}

Figure \ref{fig1} shows that the loss values of models incorporating the multi-order graph attention mechanism are significantly lower than those without this mechanism under all conditions. These results not only validate the effectiveness of the multi-order attention mechanism in enhancing predictive accuracy but also highlight that the performance gains are especially significant under highly sparse conditions, such as at 2.5\% density. This improvement can be attributed to the multi-level attention mechanism's ability to aggregate information from distant nodes, which is particularly beneficial in low-density matrices.

\subsubsection{Impact of Adversarial Neural Network (RQ3)}

Modeling user–service interactions is a critical component of our architecture, as it enables the network to effectively learn complex relational patterns. To enhance the modeling of these interactions, we incorporated adversarial neural networks into the framework. To assess their contribution, we conducted an ablation experiment by removing the adversarial component from the interaction stage. Specifically, we explored the impact of using adversarial neural networks on model results.

\begin{figure}[htbp]
    \begin{minipage}[t]{0.49\linewidth}
        \centering
        \includegraphics[width=\textwidth]{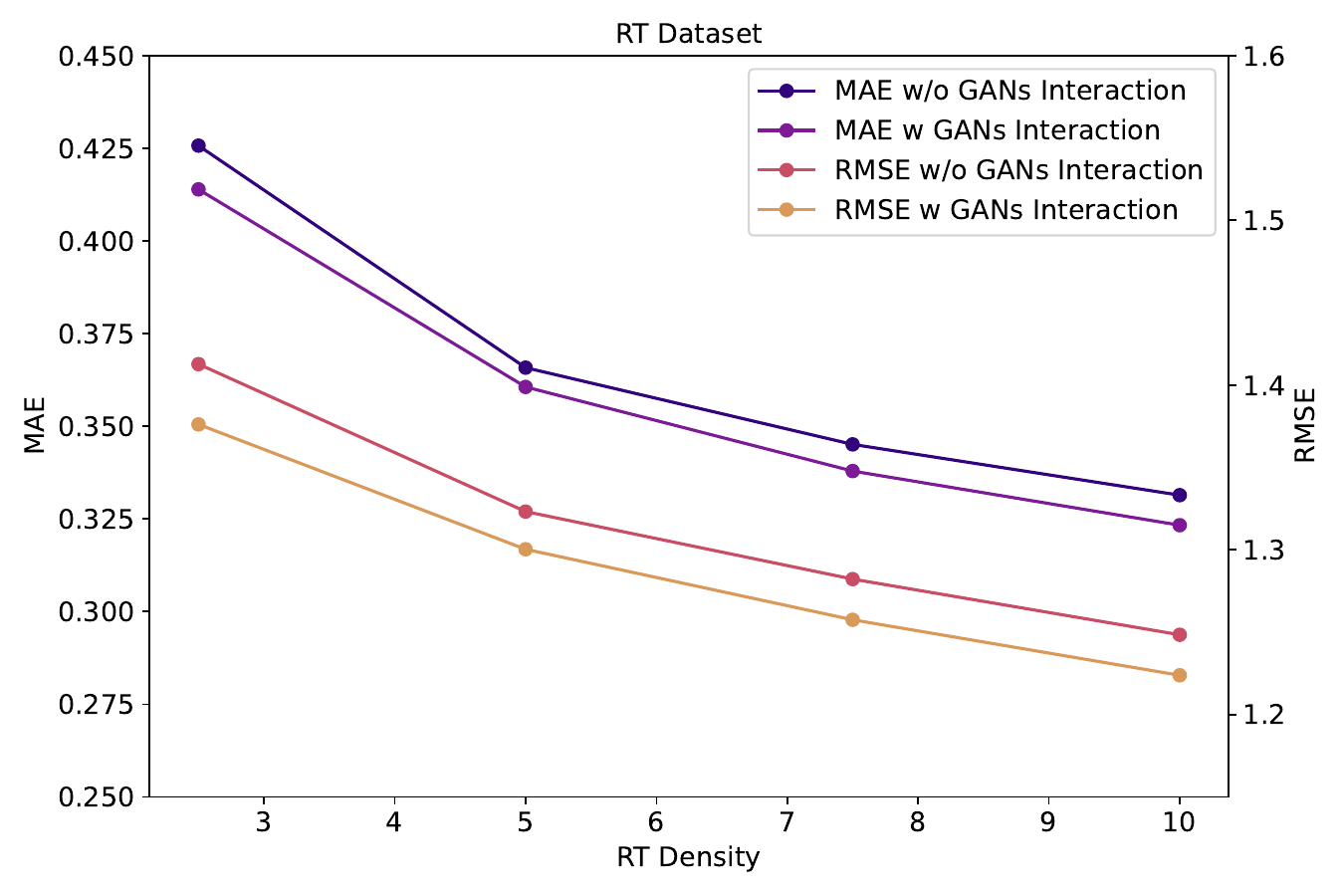}
        \centering 
    \end{minipage}
    \begin{minipage}[t]{0.49\linewidth}
        \centering
        \includegraphics[width=\textwidth]{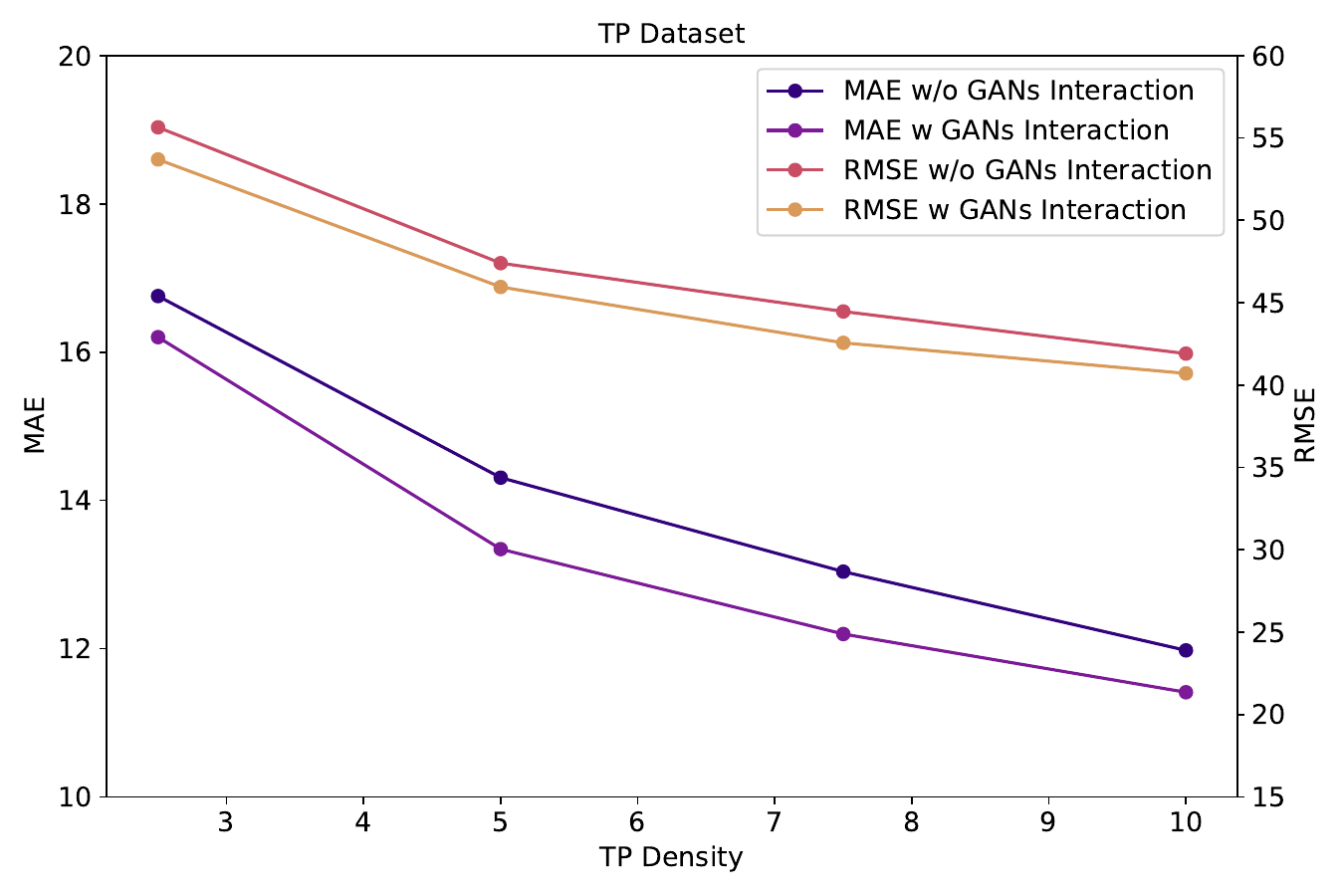}
        \centering  
    \end{minipage}
    \\
    \begin{minipage}[t]{0.49\linewidth}
    \centering
    \includegraphics[width=\textwidth]{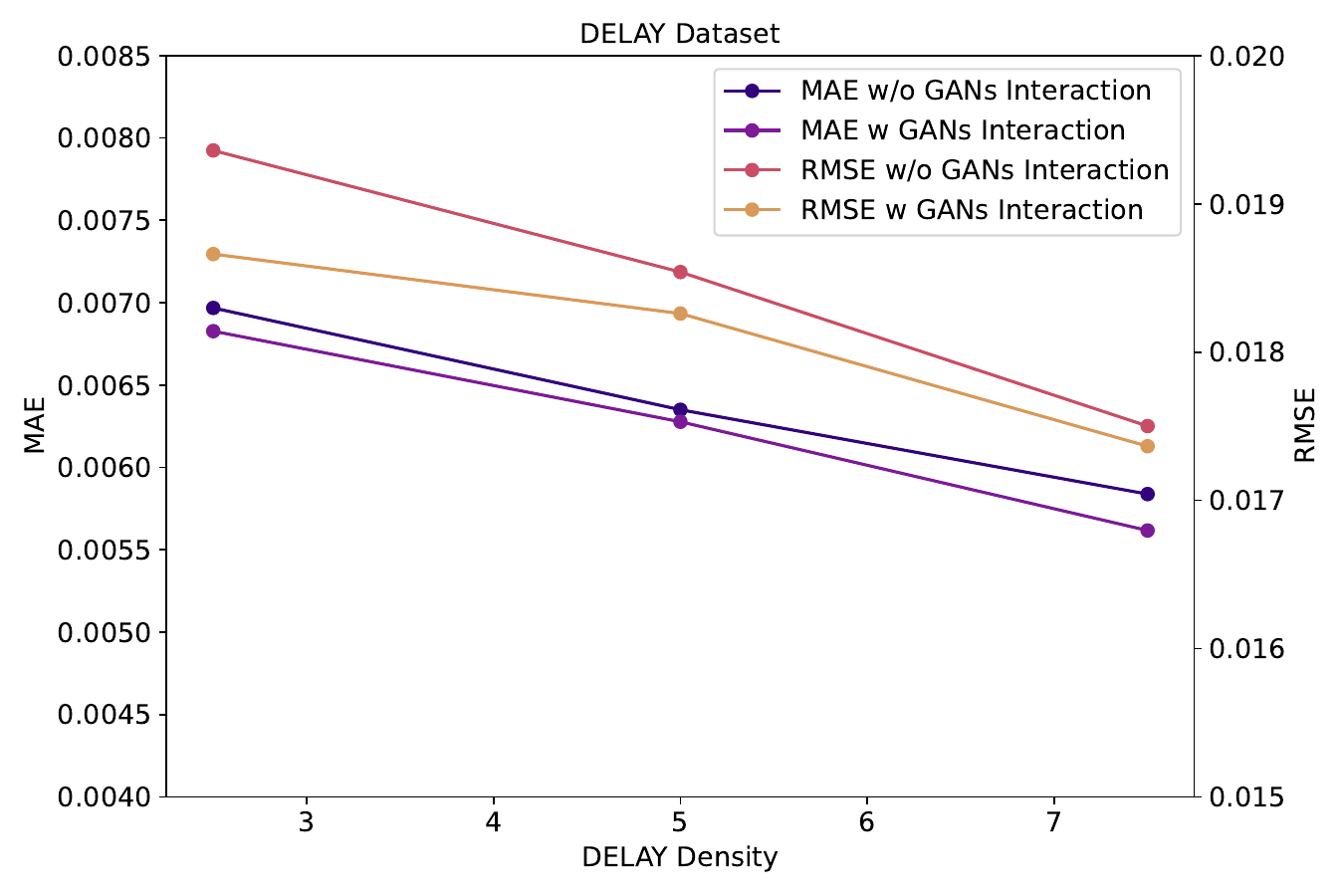}
    \centering   
\end{minipage}
\begin{minipage}[t]{0.49\linewidth}
    \centering
    \includegraphics[width=\textwidth]{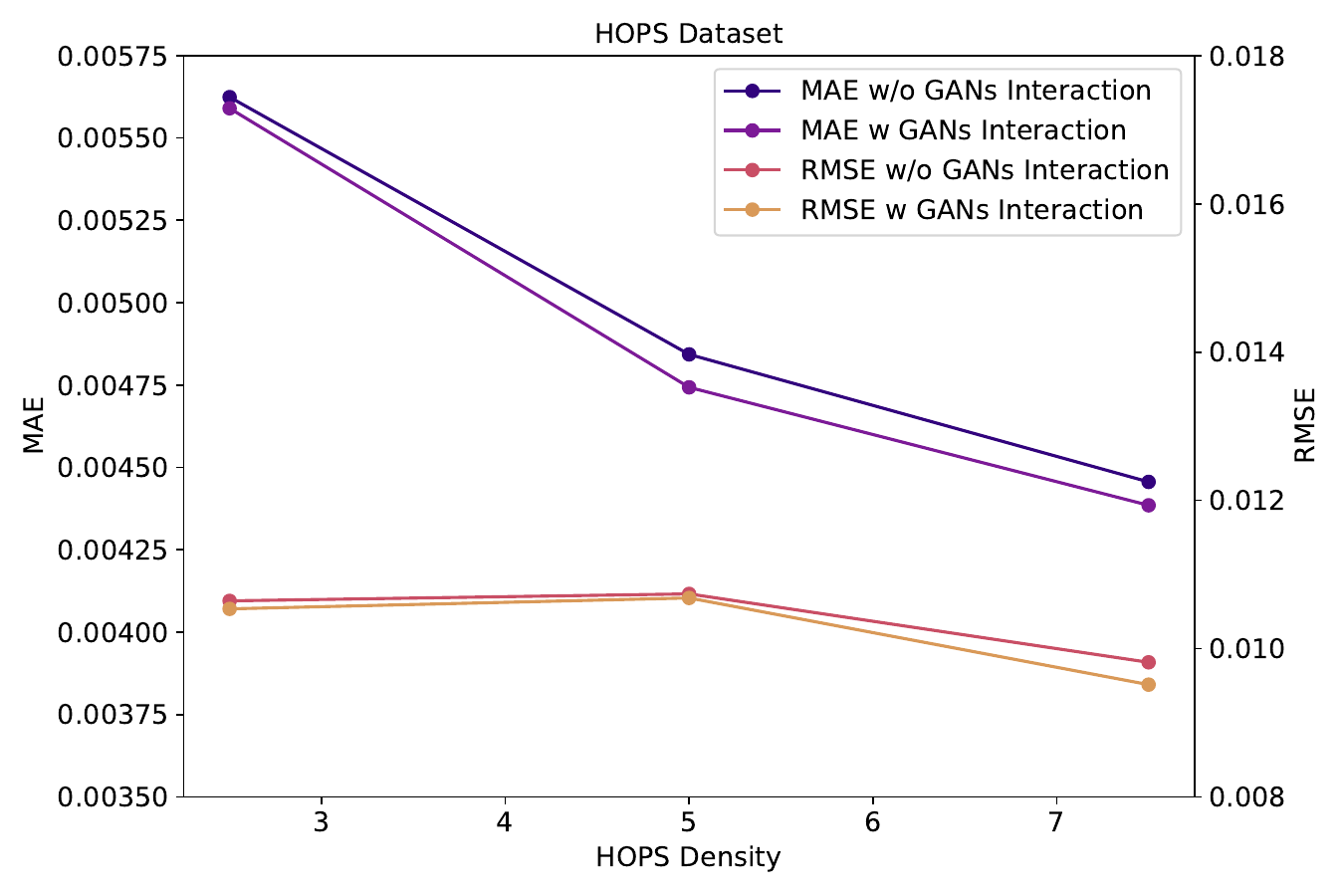}
    \centering   
\end{minipage}

    \caption{Ablation study on Adversarial Neural Networks}
    \label{fig2}
\end{figure}


As shown in the results in Figure \ref{fig2}, incorporating adversarial neural networks significantly improves prediction accuracy during the interaction modeling phase. This proves the effectiveness of our self-regression model learning.

\subsubsection{Impact of Negative Sampling on Discrete Data (RQ4)}

Generating negative samples is essential for training adversarial networks.
However, to demonstrate that traditional continuous negative sampling strategies are worse for discrete data, we conducted an ablation experiment comparing different sampling approaches.
Specifically, we compared two sampling strategies: conventional continuous sampling and Gumbel-Softmax-based discrete sampling.

\begin{table}[h]
\caption{Impact of Negative Sampling on Discrete Data in RT Datasets using different matrix densities}
\resizebox{\textwidth}{!}{
\tiny
\begin{threeparttable}
\begin{tabular}{c|cc|cc|cc|cc}
\hline
\multirow{2}{*}{\diagbox{\textbf{Methods}}{\textbf{Density}}} & \multicolumn{2}{c|}{\textbf{MD=2.5\%}} & \multicolumn{2}{c|}{\textbf{MD=5\%}} & \multicolumn{2}{c|}{\textbf{MD=7.5\%}} & \multicolumn{2}{c}{\textbf{MD=10\%}} \\ \cline{2-9} 
                                  & \textbf{MAE}       & \textbf{RMSE}      & \textbf{MAE}       & \textbf{RMSE}      & \textbf{MAE}       & \textbf{RMSE}      & \textbf{MAE}       & \textbf{RMSE}      \\ \hline

\textbf{w/o Gumbel} & 0.4139 & 1.3571 &  0.3606& 1.3003 & 0.3379, &  1.2675 & 0.3239 &  1.2233\\ \hline
\textbf{w Gumbel} & {0.397} & {1.388}     & {0.349}     & {1.291}     & {0.331}     & {0.264} & {0.316}     & {1.230}     \\
\hline
\end{tabular}
\label{rtGB}
\end{threeparttable}
}
\label{tab_rst}
\end{table}

\begin{table}[h]
\caption{Impact of Negative Sampling on Discrete Data in TP Datasets using different matrix densities}
\resizebox{\textwidth}{!}{
\tiny
\begin{threeparttable}
\begin{tabular}{c|cc|cc|cc|cc}
\hline
\multirow{2}{*}{\diagbox{\textbf{Methods}}{\textbf{Density}}} & \multicolumn{2}{c|}{\textbf{MD=2.5\%}} & \multicolumn{2}{c|}{\textbf{MD=5\%}} & \multicolumn{2}{c|}{\textbf{MD=7.5\%}} & \multicolumn{2}{c}{\textbf{MD=10\%}} \\ \cline{2-9} 
                                  & \textbf{MAE}       & \textbf{RMSE}      & \textbf{MAE}       & \textbf{RMSE}      & \textbf{MAE}       & \textbf{RMSE}      & \textbf{MAE}       & \textbf{RMSE}      \\ \hline

\textbf{w/o Gumble}  &16.4035& 53.9513& 13.9134 & 46.537 & 12.3758 &43.2247 & 11.5767 & 40.6566 \\ \hline
\textbf{w Gumbel} & {15.75} & {53.182}     & {12.988}     & {45.539}     & {11.694}     & {41.888} & {11.090}     & {39.856}     \\
\hline
\end{tabular}
\label{tpGB}
\end{threeparttable}
}
\label{tab_rst}
\end{table}

\begin{table}[h]
\caption{Impact of Negative Sampling on Discrete Data in DELAY Datasets using different matrix densities}
\resizebox{\textwidth}{!}{
\tiny
\begin{threeparttable}
\begin{tabular}{c|cc|cc|cc}
\hline
\multirow{2}{*}{\diagbox{\textbf{Methods}}{\textbf{Density}}} & \multicolumn{2}{c|}{\textbf{MD=2.5\%}} & \multicolumn{2}{c|}{\textbf{MD=5\%}} & \multicolumn{2}{c}{\textbf{MD=7.5\%}} \\ \cline{2-7} 
                                  & \textbf{MAE}       & \textbf{RMSE}      & \textbf{MAE}       & \textbf{RMSE}      & \textbf{MAE}       & \textbf{RMSE}      \\ \hline

\textbf{w/o Gumble}  & 0.007252 & 0.019168 & 0.006767 & 0.018894 & 0.006121 & 0.016864 \\ \hline
\textbf{w Gumbel}    & 0.007199 &0.019158 & 0.006588 & 0.018664 & 0.005882 & 0.017478 \\
\hline
\end{tabular}
\label{deGB}
\end{threeparttable}
}
\label{tab_rst}
\end{table}

\begin{table}[h]
\caption{Impact of Negative Sampling on Discrete Data in HOPS Datasets using different matrix densities}
\resizebox{\textwidth}{!}{
\tiny
\begin{threeparttable}
\begin{tabular}{c|cc|cc|cc}
\hline
\multirow{2}{*}{\diagbox{\textbf{Methods}}{\textbf{Density}}} & \multicolumn{2}{c|}{\textbf{MD=2.5\%}} & \multicolumn{2}{c|}{\textbf{MD=5\%}} & \multicolumn{2}{c}{\textbf{MD=7.5\%}} \\ \cline{2-7} 
                                  & \textbf{MAE}       & \textbf{RMSE}      & \textbf{MAE}       & \textbf{RMSE}      & \textbf{MAE}       & \textbf{RMSE}      \\ \hline

\textbf{w/o Gumble}  & 0.005735 & 0.010943 & 0.00482 & 0.010622 & 0.005012 & 0.009975 \\ \hline
\textbf{w Gumbel}    & 0.005590 & 0.010536 &0.004743 & 0.010683 & 0.004385 & 0.009514 \\
\hline
\end{tabular}
\label{hopsGB}
\end{threeparttable}
}
\label{tab_rst}
\end{table}

As shown in Tables~\ref{rtGB}–\ref{hopsGB}, models employing Gumbel-Softmax-based sampling consistently achieved better predictive performance and lower error metrics across all test settings. Notably,  the performance improvements brought by Gumbel-Softmax were most prominent under extremely sparse conditions (e.g., 2.5\% density), compared to higher densities such as 10\%.
This observation can be attributed to the fact that sparser matrices exhibit higher data discreteness, for which Gumbel-Softmax is particularly well-suited.

\subsection{Hyperparameter Exploration(RQ5)}

\subsubsection{Multi-order Graph Attention Layers}

\begin{figure}[htbp]
    \centering
    \includegraphics[width=\textwidth]{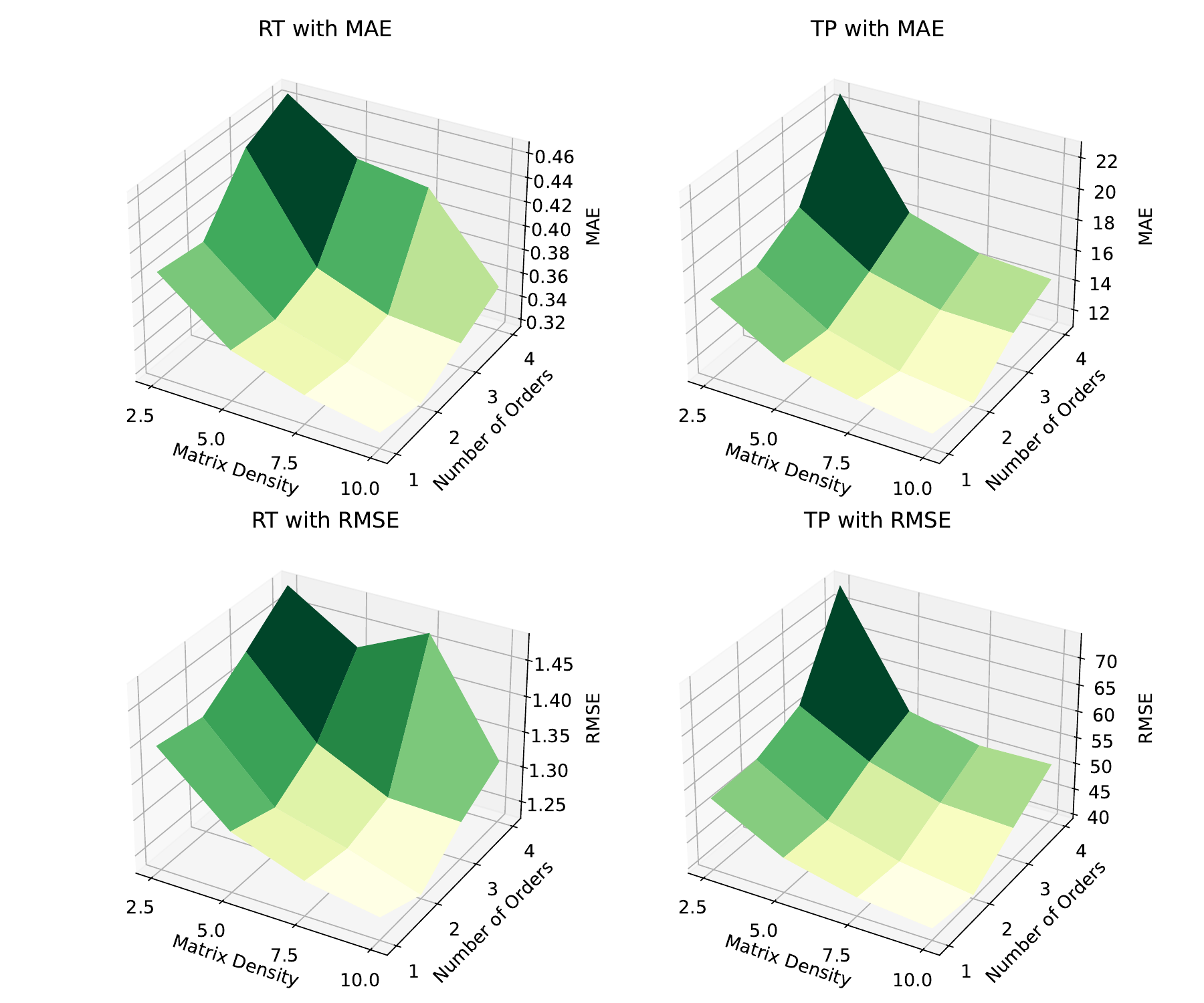}  
    \caption{Impact of Attention Orders in the Multi-Order Graph Attention Mechanism}

    \label{order}
\end{figure}

The number of orders in GAT refers to the number of steps in a GAT, during which information flows from a node to its higher-order neighbors. Such a multi-hop aggregation strategy enables the model to capture higher-order structural dependencies in the graph. To explore the impact of the attention order on our model, we conducted experiments on different datasets with varying attention layer depths, as shown in Figures \ref{order}, using MAE and RMSE as performance indicators.

The experimental results reveal similar performance trends across both RT and TP datasets: increasing the attention order from one to two significantly enhances model performance. However, further increasing the order beyond two leads to a performance degradation. This suggests that incorporating up to second-order neighbors is sufficient for achieving optimal performance in the context of QoS prediction. Notably, although second-order attention continues to benefit the model under higher graph densities, the performance gains are less significant compared to those under sparse conditions. Therefore, we set the attention order to 2 in all subsequent main experiments.

\subsubsection{Impact of Generator Balance Loss Parameter $\lambda$}

\begin{figure}[htbp]
    \centering
    \includegraphics[width=\textwidth]{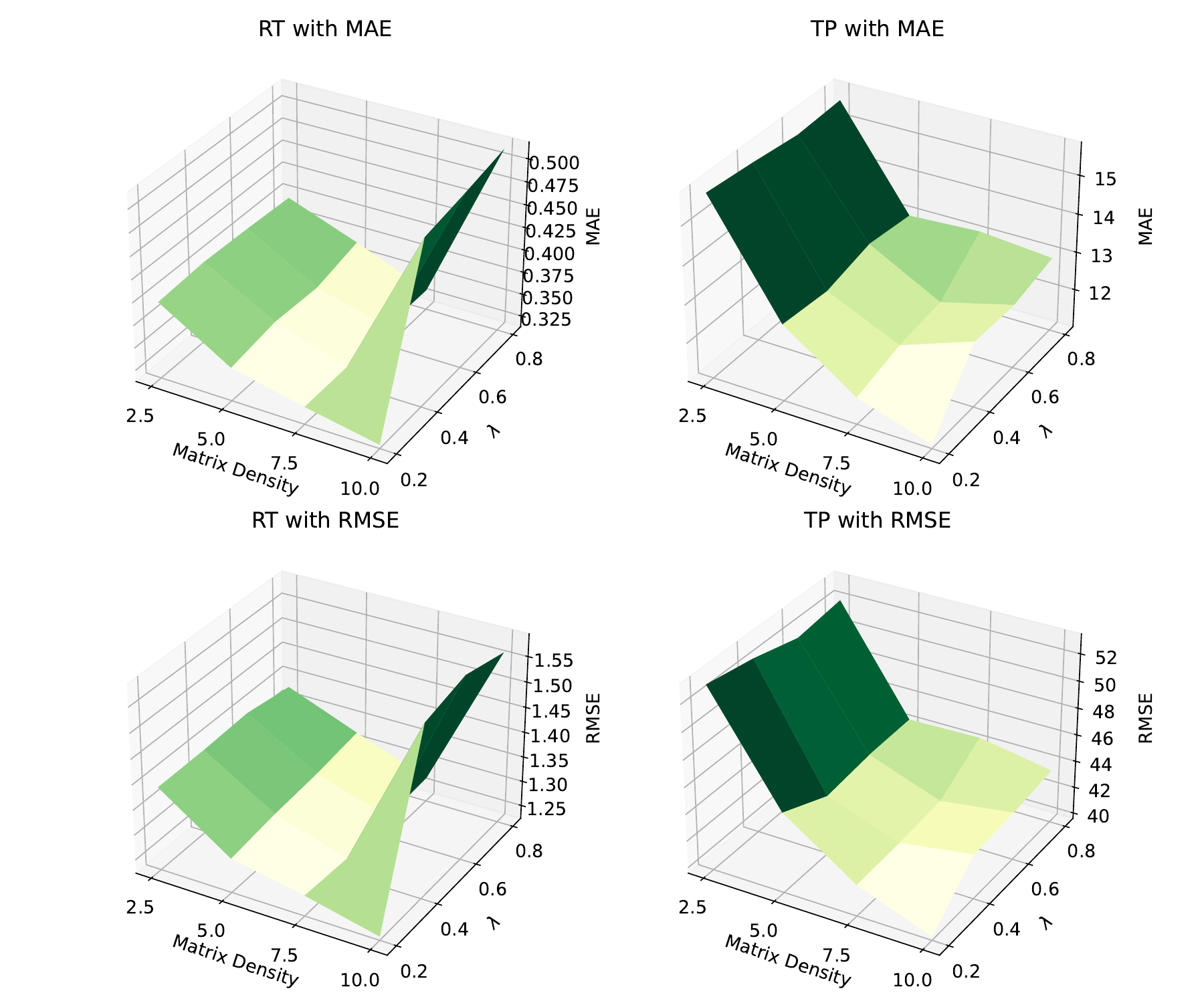}  
    \caption{Balance Parameter $\lambda$ Exploration}
    \label{lmd}
\end{figure}

In our model, the balance loss parameter $\lambda$ controls the trade-off between two components: the binary classification loss of the discriminator on real samples and the regression loss between the generator’s output and the ground truth. It is used to regulate the generator’s training dynamics within the adversarial framework.
This parameter $\lambda$ was specifically introduced to explore the weighting relationship between these two losses to ensure the effectiveness of the autoregressive process. As illustrated in Figure~\ref{lmd}, we evaluated the impact of varying $\lambda$ values on model performance.

The results indicate a general performance degradation with increasing values of $\lambda$.
This suggests that relatively smaller $\lambda$ values are more appropriate for our model architecture. Although the discriminator's output is integral to the autoregressive framework, assigning it excessive weight may compromise the model's stability and overall performance. Based on the above findings, we set $\lambda = 0.2$ in the main experiments.

\subsubsection{Impact of Negative Sampling Temperature}

\begin{figure}[htbp]
    \centering
    \includegraphics[width=\textwidth]{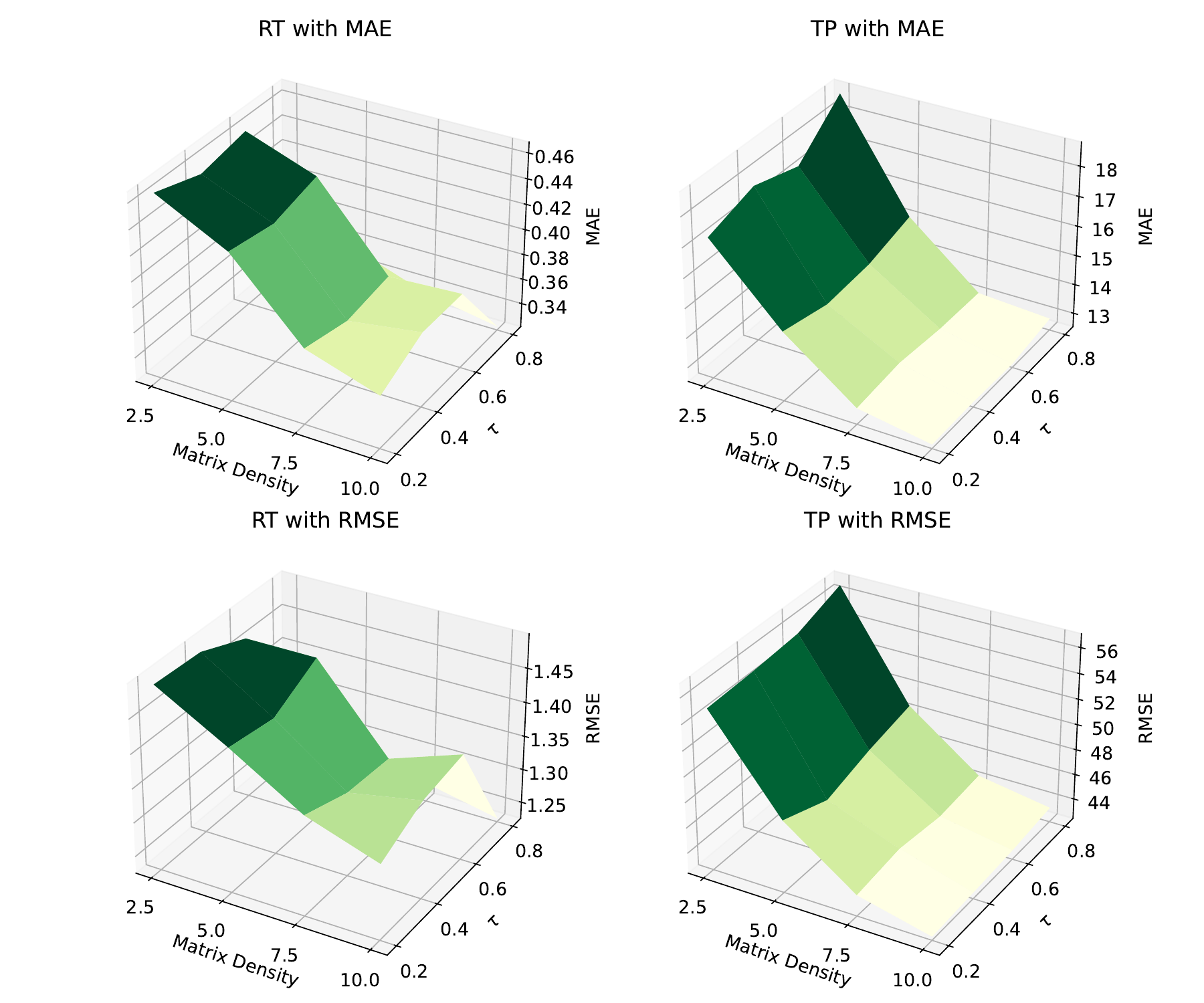}  
    \caption{Exploration of Temperature Parameter $\tau$ in Gumbel-Softmax}
    \label{tau}
\end{figure}

 The temperature parameter $tau$ in  Gumbel-Softmax plays a role in adjusting the "hardness" of the output sampling distribution. Specifically, when the temperature $tau$ is low, the output tends to hard sampling, meaning the sampling results are closer to discrete 0 and 1, which helps ensure clearer decision boundaries in the later stages of model training. In contrast, when the temperature $tau$ is high, the output exhibits softer characteristics, with the sampling distribution being more uniform and the probabilities of each category being closer. This property facilitates exploratory learning during the early stages of training and maintains smooth gradients, preventing the model from getting stuck in local optima prematurely. During the hyperparameter optimization of the model, we evaluated the effect of different temperature parameters $t$ on model performance through experiments to determine the optimal setting. The experimental results are shown in Figures \ref{tau}.

The results confirm that the temperature parameter $\tau$ influences model performance, though its effect varies across datasets and matrix density levels. We attribute this variation to the differing levels of sparsity and distributional characteristics across datasets, which affect the optimal sampling behavior. Based on these observations, we selected $\tau = 0.5$ as the temperature parameter for all main experiments.

\subsection{Inference Time and Memory Analysis(RQ6)}

\subsubsection{Inference time}Inference time is a critical factor in QoS prediction tasks, particularly in latency-sensitive scenarios such as real-time service selection and adaptive routing. In such applications, timely decisions directly affect user experience and overall system performance. Therefore, we measured the inference time of our model across four datasets under varying matrix sparsity levels, where each inference corresponds to a single user–service interaction.

The results are summarized in Table~\ref{tab:inference_time}.

\begin{table}[htbp]
\centering
\caption{Inference Time (seconds) under Different Matrix Densities}
\label{tab:inference_time}
\begin{tabular}{c|cccc}
\hline
\textbf{Dataset} & \textbf{2.5\%} & \textbf{5\%} & \textbf{7.5\%} & \textbf{10\%} \\
\hline
RT     & 0.00363 & 0.00392 & 0.00370 & 0.00446 \\
TP     & 0.00354 & 0.00342 & 0.00350 & 0.00352 \\
DELAY  & 0.00219 & 0.00225 & 0.00271 & 0.00257 \\
HOPS   & 0.00210 & 0.00227 & 0.00219 & 0.00357 \\
\hline
\end{tabular}
\end{table}

Among the datasets, the inference time on the RT and TP datasets remains relatively stable across density levels, with a slight increase observed as density grows. This trend is expected since denser matrices involve more observed interactions, thereby increasing the computational overhead during inference. Notably, the maximum inference latency on the RT dataset at 10\% density reaches 0.00446 seconds, which remains within an acceptable range for most latency-critical systems.

In contrast, the DELAY and HOPS datasets exhibit slightly lower inference times overall, particularly at lower densities (2.5\% and 5\%). This may be attributed to the relatively simpler graph structures in these datasets, enabling faster model inference. However, an outlier is observed on the HOPS dataset at 10
 \% density, where the inference time peaks at 0.00357 seconds, higher than its lower-density counterparts. This suggests that increasing matrix density, even in structurally simpler datasets, can lead to non-linear growth in computational cost due to the heightened complexity of interaction modeling.

\subsubsection{Memory Analysis}

\begin{figure}[htbp]
    \centerline{\includegraphics[width=0.98\textwidth]{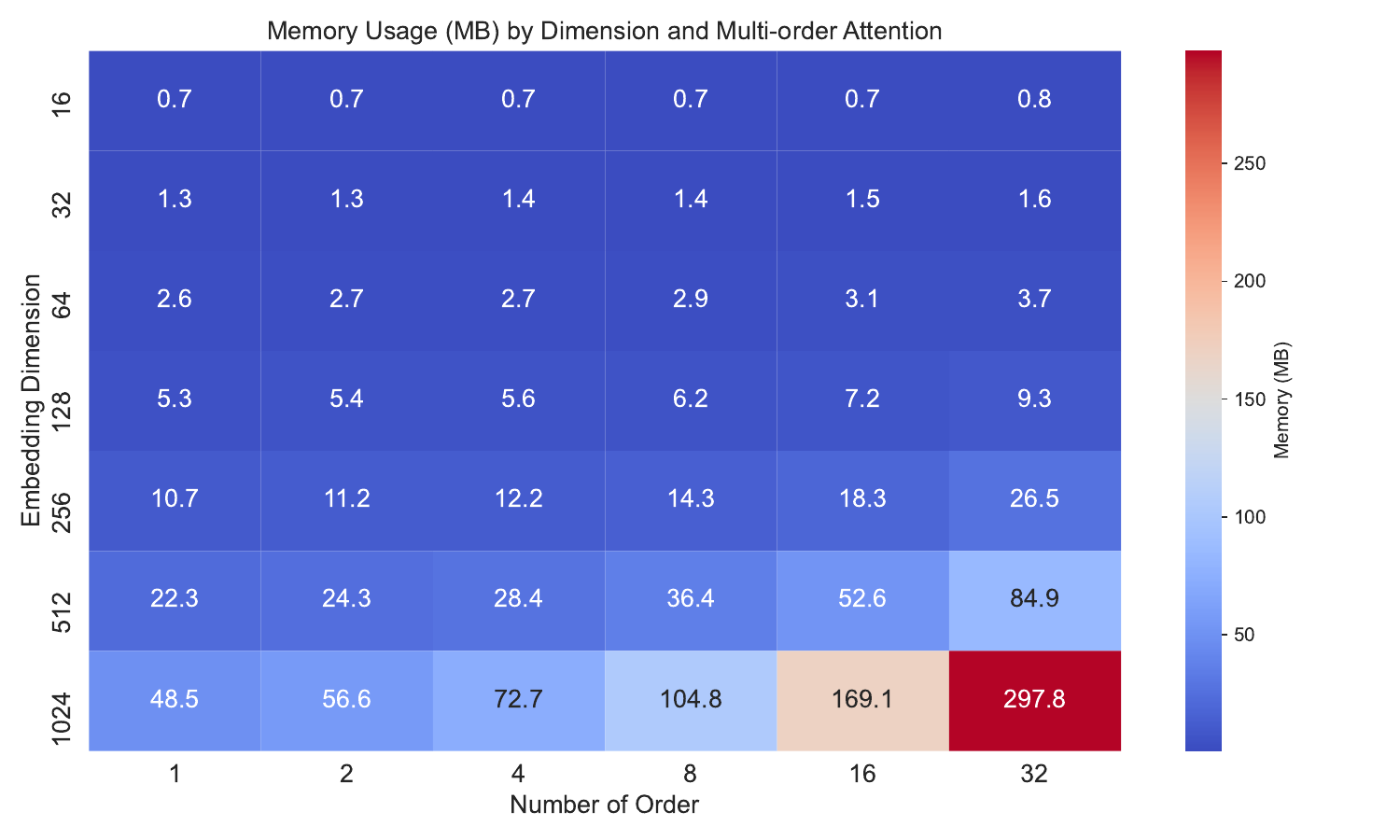}}
    \caption{Memory Usage of QoSMGAA}  
    \label{memory} 
\end{figure}

Evaluating memory overhead is crucial in QoS. Memory overhead is another critical consideration in QoS prediction tasks, especially in resource-constrained environments with high concurrency requirements. A model with lower memory consumption enables better scalability, real-time responsiveness, and broader deployment feasibility in large-scale service systems. We evaluated the memory overhead required by models with different hyperparameter settings based on the QoSGMAA framework, and the results are shown in Figure \ref{memory}.

As illustrated in Figure~\ref{memory}, the memory consumption of the QoSGMAA framework increases progressively with both the embedding dimension and the number of orders in Multi-Order GAT. Specifically, doubling the embedding dimension leads to a near-quadratic increase in memory usage, which is further exacerbated by the rise in the number of attention orders. For example, increasing the embedding dimension from 128 to 256 (with 16 attention heads) results in a substantial memory increase from approximately 7.2 MB to 18.3 MB. Furthermore, increasing the number of attention heads from 8 to 32 at a fixed embedding dimension of 1024 causes memory usage to surge from 104.8 MB to nearly 298 MB, highlighting the compounding resource demands introduced by multi-order attention mechanisms.

To balance predictive performance and computational efficiency, we selected an embedding dimension of 32 and set the number of attention orders to 2, corresponding to a second-order multi-head attention configuration. Under this setting, the memory overhead remains relatively low—approximately 1.35 MB—making it highly suitable for deployment in server environments where scalable and efficient QoS prediction is required. 
These findings demonstrate the practical value of the proposed model in real-world service systems with limited computational resources.

\subsection{Model Robustness Test (RQ7)}

Robustness analysis plays a vital role in QoS prediction tasks, as it directly reflects a model's ability to maintain consistent performance under realistic and potentially adverse conditions. In practical deployment scenarios, QoS data often suffers from various forms of noise, including measurement errors, incomplete interactions, and dynamic environmental fluctuations. These imperfections can significantly degrade the reliability of prediction models overly sensitive to input variations. 

In this section, we investigate the robustness of the proposed QoSGMAA framework. To evaluate the model's performance in the presence of varying noise levels, we introduce synthetic structural noise during the graph construction phase by randomly replacing a certain percentage of true edges with randomly generated false edges. The noisy graphs are then used for training, validation, and testing.

\begin{figure}[htbp]
    \begin{minipage}[t]{\linewidth}
        \centering
        \includegraphics[width=\textwidth]{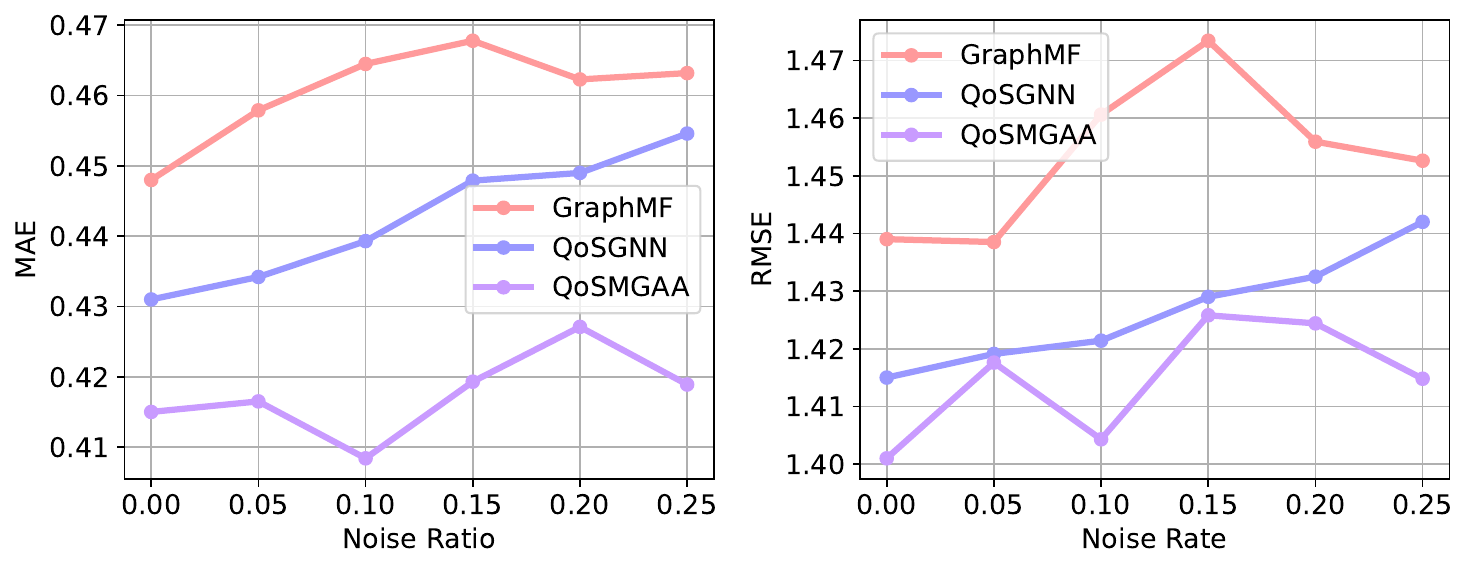}
        \centering  (a) RT dataset 
    \end{minipage}
    \\
    \begin{minipage}[t]{\linewidth}
        \centering
        \includegraphics[width=\textwidth]{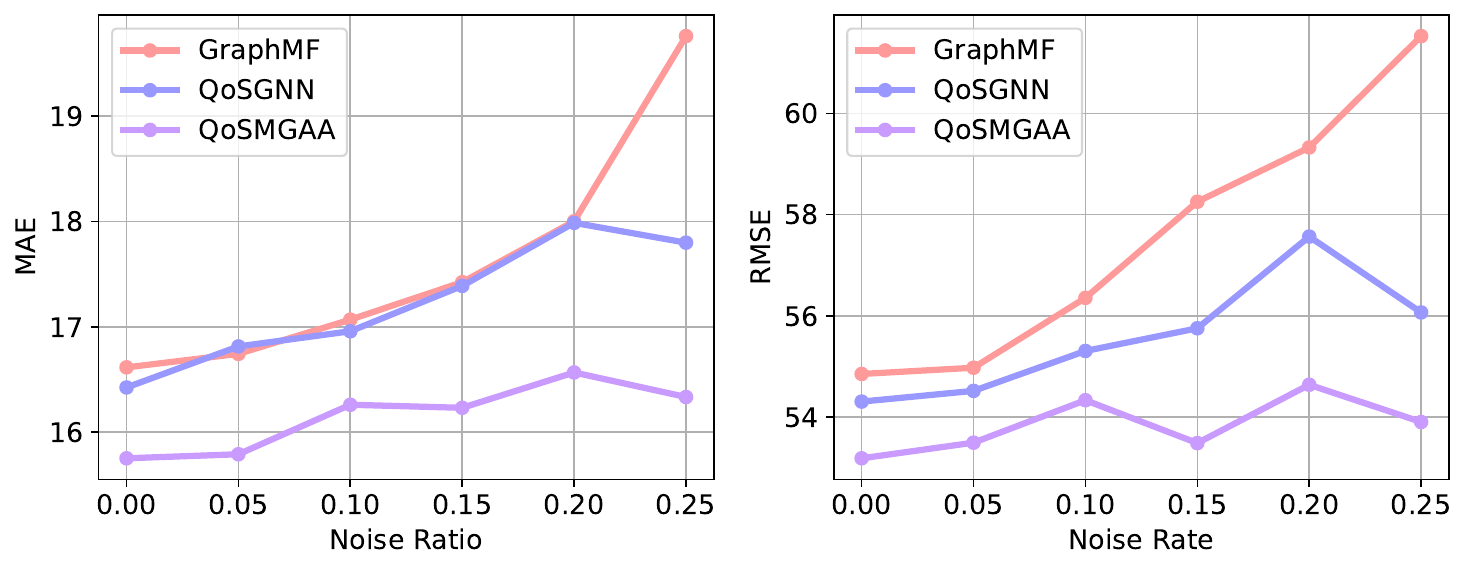}
        \centering  (b) TP dataset 
    \end{minipage}
    \caption{Relative performance degradation w.r.t noise ratio.}
    \label{figrob}
\end{figure}

Specifically, we conduct experiments on the RT and TP datasets under a fixed matrix sparsity level of 2.5\%. For each dataset, we simulate five different noise levels by replacing 5\%, 10\%, 15\%, 20\%, and 25\% of the original edges with noise. We compare the performance of QoSGMAA with two representative graph neural network-based baselines: QoSGNN and GraphMF. The results, presented in Figure \ref{figrob},

On the RT dataset, QoSMGAA maintained remarkable stability in both MAE and RMSE as the noise ratio increased from 0\% to 25\%. The MAE of QoSMGAA fluctuated only slightly from 0.415 to 0.4189, corresponding to a relative increase of merely 0.94\%. In contrast, GraphMF’s MAE rose from 0.448 to 0.4632, a 3.39\% increase, and QoSGNN’s MAE increased by 5.48\%, from 0.431 to 0.4546. Similarly, QoSMGAA’s RMSE rose by less than 1\% (from 1.401 to 1.4148), while GraphMF and QoSGNN saw increases of 0.95\% and 1.91\%, respectively. This demonstrates that QoSMGAA not only achieves lower absolute prediction errors but also resists performance degradation more effectively under noisy conditions.

A similar pattern emerged on the TP dataset. Although the absolute MAE and RMSE values were higher due to the dataset’s larger scale, the relative stability of QoSMGAA remained evident. The MAE of QoSMGAA increased modestly from 15.750 to 16.3323 (a 3.69\% rise), whereas GraphMF's MAE escalated by 18.92\% and QoSGNN’s by 8.37\%. Regarding RMSE, QoSMGAA’s increase was only 1.35\% (from 53.182 to 53.9), significantly smaller than GraphMF’s 12.18\% and QoSGNN’s 3.25\%. These results indicate that QoSMGAA maintains superior accuracy and provides enhanced robustness to structural distortions across different datasets.

This robustness can be attributed to the model’s architectural innovations. The multi-order graph attention mechanism allows QoSMGAA to capture global structural dependencies, thereby reducing the model’s reliance on any individual edge and mitigating the impact of random edge perturbations. Furthermore, the adversarial training strategy equips the model with the ability to discriminate between true and noisy patterns, improving its generalization under uncertain graph structures. These findings underscore QoSMGAA’s effectiveness in maintaining prediction reliability in noisy and dynamic QoS environments, distinguishing it from less resilient models like GraphMF and QoSGNN not only in terms of performance but also in terms of robustness.


%% file: text/rt_new_table.tex
\begin{table}[]
\centering
\caption{Performance Comparison of QoS Prediction Models on Response Time}
\label{tab:rt_comparison}
\resizebox{\textwidth}{!}{
\begin{tabular}{lcccccccc}
\hline
\multirow{2}{*}{\textbf{Model}} & \multicolumn{2}{c}{\textbf{Density=2.5\%}} & \multicolumn{2}{c}{\textbf{Density=5\%}} & \multicolumn{2}{c}{\textbf{Density=7.5\%}} & \multicolumn{2}{c}{\textbf{Density=10\%}} \\
\cline{2-9}
 & \textbf{MAE} & \textbf{RMSE} & \textbf{MAE} & \textbf{RMSE} & \textbf{MAE} & \textbf{RMSE} & \textbf{MAE} & \textbf{RMSE} \\
\hline
UPCC    & 0.709$\pm$0.007 & 1.467$\pm$0.008 & 0.640$\pm$0.021 & 1.380$\pm$0.003 & 0.588$\pm$0.003 & 1.339$\pm$0.003 & 0.556$\pm$0.003 & 1.309$\pm$0.004\\
IPCC    & 0.755$\pm$0.006 & 1.657$\pm$0.004 & 0.637$\pm$0.001 & 1.399$\pm$0.003 & 0.615$\pm$0.002 & 1.367$\pm$0.002 & 0.596$\pm$0.002 & 1.343$\pm$0.002\\
UIPCC   & 0.737$\pm$0.006 & 1.615$\pm$0.005 & 0.628$\pm$0.001 & 1.388$\pm$0.003 & 0.604$\pm$0.002 & 1.355$\pm$0.002 & 0.584$\pm$0.002 & 1.330$\pm$0.002\\
PMF     & 0.712$\pm$0.004 & 1.842$\pm$0.007 & 0.570$\pm$0.001 & 1.535$\pm$0.005 & 0.561$\pm$0.004 & 1.395$\pm$0.004 & 0.487$\pm$0.002 & 1.313$\pm$0.004\\
BiasMF  & 0.689$\pm$0.002 & 1.537$\pm$0.008 & 0.600$\pm$0.014 & 1.385$\pm$0.004 & 0.544$\pm$0.002 & 1.311$\pm$0.004 & 0.516$\pm$0.003 & 1.263$\pm$0.005\\
CSMF    & 0.649$\pm$0.003 & 1.678$\pm$0.009 & 0.550$\pm$0.001 & 1.494$\pm$0.002 & 0.497$\pm$0.002 & 1.407$\pm$0.003 & 0.453$\pm$0.001 & 1.356$\pm$0.001\\
NFMF    & 0.524$\pm$0.004 & 1.484$\pm$0.031 & 0.447$\pm$0.005 & 1.355$\pm$0.010 & 0.426$\pm$0.001 & 1.322$\pm$0.006 & 0.413$\pm$0.003 & 1.303$\pm$0.004\\
NCRL    & 0.561$\pm$0.005 & 1.591$\pm$0.019 & 0.546$\pm$0.002 & 1.564$\pm$0.016 & 0.542$\pm$0.004 & 1.540$\pm$0.015 & 0.537$\pm$0.004 & 1.519$\pm$0.007\\
GraphMF & 0.448$\pm$0.005 & 1.415$\pm$0.011 & 0.399$\pm$0.007 & 1.345$\pm$0.012 & 0.378$\pm$0.005 & 1.302$\pm$0.009 & 0.367$\pm$0.003 & 1.299$\pm$0.008\\
QoSGNN  & 0.431$\pm$0.005 & 1.439$\pm$0.009 & 0.377$\pm$0.006 & 1.335$\pm$0.008 & 0.353$\pm$0.004 & 1.295$\pm$0.011 & 0.345$\pm$0.004 & 1.276$\pm$0.010\\ \hline
QoSMGAA & 0.397$\pm$0.005 & 1.388$\pm$0.001 & 0.349$\pm$0.004 & 1.291$\pm$0.001 & 0.331$\pm$0.004 & 1.264$\pm$0.005 & 0.316$\pm$0.00 & 1.230$\pm$0.002\\
\textit{imp.} & \textit{16.33\%} & \textit{7.84\%} & \textit{10.32\%} & \textit{3.42\%} & \textit{20.102\%} & \textit{7.35\%} & \textit{16.00\%} & \textit{4.72\%}\\
\hline
\end{tabular}
}
\end{table}

%% file: text/tp_new_table.tex
\begin{table}[]
\centering
\caption{Performance Comparison of QoS Prediction Models on Throughput}
\label{tab:tp_comparison}
\resizebox{\textwidth}{!}{
\begin{tabular}{lcccccccc}
\hline
\multirow{2}{*}{\textbf{Model}} & \multicolumn{2}{c}{\textbf{Density=2.5\%}} & \multicolumn{2}{c}{\textbf{Density=5\%}} & \multicolumn{2}{c}{\textbf{Density=7.5\%}} & \multicolumn{2}{c}{\textbf{Density=10\%}} \\
\cline{2-9}
 & \textbf{MAE} & \textbf{RMSE} & \textbf{MAE} & \textbf{RMSE} & \textbf{MAE} & \textbf{RMSE} & \textbf{MAE} & \textbf{RMSE} \\
\hline
UPCC   & $31.759\pm0.166$ & $67.816\pm0.221$ & $27.209\pm0.159$ & $60.907\pm0.179$ & $24.4925\pm0.126$ & $57.176\pm0.039$ & $22.605\pm0.085$ & $54.522\pm0.113$ \\
IPCC   & $31.781\pm0.283$ & $73.172\pm0.515$ & $27.0769\pm0.129$ & $62.929\pm0.126$ & $26.415\pm0.079$ & $61.318\pm0.067$ & $26.182\pm0.053$ & $60.353\pm0.061$ \\
UIPCC  & $31.260\pm0.160$ & $67.397\pm0.221$ & $26.749\pm0.143$ & $60.683\pm0.175$ & $24.166\pm0.118$ & $57.041\pm0.043$ & $22.364\pm0.077$ & $54.421\pm0.112$ \\
PMF    & $24.287\pm0.280$ & $72.125\pm0.844$ & $19.078\pm0.147$ & $57.814\pm0.508$ & $17.024\pm0.135$ & $51.516\pm0.423$ & $16.027\pm0.073$ & $48.256\pm0.246$ \\
BiasMF & $28.618\pm0.320$ & $69.494\pm0.861$ & $21.816\pm0.195$ & $56.766\pm0.510$ & $19.260\pm0.108$ & $51.277\pm0.296$ & $17.883\pm0.139$ & $48.390\pm0.263$ \\
CSMF & $26.1575\pm0.254$ & $ 72.285\pm 0.584$ & $20.824\pm0.085$ & $58.806\pm0.586$ & $18.236\pm0.082$ & $52.563\pm0.374$ & $16.691\pm0.062$ & $48.996\pm0.354$ \\
NFMF   & $25.878\pm0.289$ & $67.975\pm1.737$ & $20.086\pm0.307$ & $54.626\pm0.487$ & $18.252\pm0.368$ & $51.802\pm0.957$ & $17.602\pm0.222$ & $50.900\pm0.970$ \\
NCRL   & $28.418\pm0.443$ & $80.555\pm1.702$ & $27.130\pm0.414$ & $77.011\pm0.785$ & $26.867\pm0.298$ & $75.953\pm1.127$ & $26.648\pm0.230$ & $75.413\pm0.502$ \\
GraphMF & $19.514\pm0.745$ & $59.849\pm1.093$ & \textbf{$16.140\pm1.100$} & \textbf{$52.489\pm2.621$} & $15.648\pm0.901$ & $50.953\pm2.182$ & $15.287\pm0.833$ & $50.171\pm2.322$ \\
QoSGNN & $18.823\pm0.045$ & $59.303\pm0.190$ & $16.255\pm0.116$ & $54.206\pm0.188$ & $14.489\pm0.359$ & $48.640\pm1.013$ & $13.946\pm0.355$ & $47.955\pm0.756$ \\ \hline
QoSMGAA & $15.75\pm0.198$ & $53.182\pm0.165$ & $12.988\pm0.174$ & $45.539\pm0.276$ & $11.694\pm0.095$ & $41.888\pm0.236$ & $11.020\pm0.019$ & $39.165\pm0.654$ \\ 
\textit{imp.} & \textit{19.29\%} & \textit{5.98\%} & \textit{13.88\%} & \textit{2.39\%} & \textit{20.48\%} & \textit{8.20\%} & \textit{16.89\%} & \textit{3.57\%}\\\hline

\end{tabular}
}
\end{table}

%% file: text/de_new_table.tex
\begin{table}[]
\centering
\caption{Performance Comparison of QoS Prediction Models on DELAY}
\label{tab:delay_comparison_new}
\resizebox{\textwidth}{!}{%
\begin{tabular}{lcccccc}
\hline
\multirow{2}{*}{\textbf{Model}} & \multicolumn{2}{c}{\textbf{Density=2.5\%}} & \multicolumn{2}{c}{\textbf{Density=5\%}} & \multicolumn{2}{c}{\textbf{Density=7.5\%}} \\
\cline{2-7}
& \textbf{MAE} & \textbf{RMSE} & \textbf{MAE} & \textbf{RMSE} & \textbf{MAE} & \textbf{RMSE} \\
\hline
PMF      & 0.008741$\pm$0.001708 & 0.020204$\pm$0.001106 & 0.007073$\pm$0.000041 & 0.019539$\pm$0.000051 & 0.007009$\pm$0.000084 & 0.018848$\pm$0.000058 \\
CSMF     & 0.008085$\pm$0.000020 & 0.020096$\pm$0.000333 & 0.007315$\pm$0.000020 & 0.018446$\pm$0.000136 & 0.006930$\pm$0.000029 & 0.017475$\pm$0.000053 \\
NFMF     & 0.007702$\pm$0.000052 & 0.019906$\pm$0.000152 & 0.007132$\pm$0.000098 & 0.018497$\pm$0.000764 & 0.006789$\pm$0.000095 & 0.017214$\pm$0.000054 \\
NCRL     & 0.007770$\pm$0.000179 & 0.019038$\pm$0.000256 & 0.007236$\pm$0.000166 & 0.018585$\pm$0.000617 & 0.006869$\pm$0.000087 & 0.017736$\pm$0.000175 \\
GraphMF  & 0.007245$\pm$0.000102 & 0.020375$\pm$0.000586 & 0.007012$\pm$0.000135 & 0.018969$\pm$0.000239 & 0.006497$\pm$0.000192 & 0.017971$\pm$0.000383 \\
QoSGNN   & 0.007199$\pm$0.000582 & 0.019158$\pm$0.000794 & 0.006588$\pm$0.000322 & 0.018664$\pm$0.000239 & 0.005882$\pm$0.000246 & 0.017478$\pm$0.000345 \\ \hline
QoSMGAA  & 0.006827$\pm$0.000014 & 0.018618$\pm$0.000869 & 0.006277$\pm$0.000126 & 0.018265$\pm$0.000204 & 0.005617$\pm$0.000158 & 0.017365$\pm$0.000343 \\
\textit{Improve} & \textit{5.17\%} & \textit{2.59\%} & \textit{4.72\%} & \textit{2.14\%} & \textit{4.51\%} & \textit{0.65\%} \\
\hline
\end{tabular}
}
\end{table}

%% file: text/ho_new_table.tex
\begin{table}[]
\centering
\caption{Performance Comparison of QoS Prediction Models on HOPS}
\label{tab:hops_comparison_newer}
\resizebox{\textwidth}{!}{%
\begin{tabular}{lcccccc}
\hline
\multirow{2}{*}{\textbf{Model}} & \multicolumn{2}{c}{\textbf{Density=2.5\%}} & \multicolumn{2}{c}{\textbf{Density=5\%}} & \multicolumn{2}{c}{\textbf{Density=7.5\%}} \\
\cline{2-7}
& \textbf{MAE} & \textbf{RMSE} & \textbf{MAE} & \textbf{RMSE} & \textbf{MAE} & \textbf{RMSE} \\
\hline
PMF        & 0.021827$\pm$0.001262 & 0.027745$\pm$0.001305 & 0.018163$\pm$0.002771 & 0.024393$\pm$0.002592 & 0.015867$\pm$0.003417 & 0.021887$\pm$0.003155 \\
CSMF       & 0.008832$\pm$0.000083 & 0.013039$\pm$0.000068 & 0.007227$\pm$0.000545 & 0.012059$\pm$0.000401 & 0.006009$\pm$0.000042 & 0.010188$\pm$0.000020 \\
NFMF       & 0.007270$\pm$0.000041 & 0.012088$\pm$0.000258 & 0.006070$\pm$0.000059 & 0.011990$\pm$0.000124 & 0.005242$\pm$0.000049 & 0.010092$\pm$0.000183 \\
NCRL       & 0.007964$\pm$0.000123 & 0.012513$\pm$0.000100 & 0.006103$\pm$0.000136 & 0.011590$\pm$0.000076 & 0.005344$\pm$0.000093 & 0.009979$\pm$0.000045 \\
GraphMF    & 0.006679$\pm$0.000302 & 0.012050$\pm$0.000337 & 0.005644$\pm$0.000106 & 0.011812$\pm$0.000063 & 0.004921$\pm$0.009820 & 0.0100009$\pm$0.000150 \\
QoSGNN     & 0.006663$\pm$0.000482 & 0.011607$\pm$0.000222 & 0.005330$\pm$0.000161 & 0.011480$\pm$0.000057 & 0.004831$\pm$0.000235 & 0.010091$\pm$0.000064 \\ \hline
QoSMGAA & 0.005590$\pm$0.000114 & 0.010536$\pm$0.000030 & 0.004743$\pm$0.000044 & 0.010683$\pm$0.000097 & 0.004385$\pm$0.000028 & 0.009514$\pm$0.000027 \\
\textit{Improve} & \textit{16.1\%} & \textit{9.23\%} & \textit{11.01\%} & \textit{6.94\%} & \textit{9.23\%} & \textit{5.72\%} \\
\hline
\end{tabular}
}
\end{table}

%% file: text/6Con.tex
 In this paper, to better capture the higher-order relationships between users and services and enable the model to learn feature representations more effectively, we propose a novel framework named QoSMGAA. This framework includes multi-order graph attention mechanisms with adversarial neural networks to improve the efficiency and accuracy of QoS prediction. Our model improves the robustness of graph-based QoS prediction systems against noise perturbation. We also achieved better prediction results on real-world datasets. Additionally, the other experiments deepen the understanding of the model's behavior and decision-making process.

In our future work, we recognize that the current framework does not fully exploit the features of users and services, which are pivotal for advanced modeling. To address this limitation, we want to utilize context-based sampling methods, which have the potential to facilitate accurate QoS predictions under specific conditions. Furthermore, the presence of noisy data can substantially degrade QoS prediction performance. In addition to the adversarial neural network currently employed, an alternative approach could involve adopting diffusion models, which have gained considerable traction in recent years. We will use the denoising diffusion probability model in the graph learning process to help reduce the impact of noise data in the graph learning process and enhance the robustness of the model. These enhancements can significantly improve the robustness and accuracy of QoS prediction models.